%% file: main.tex
\documentclass[11pt,a4paper]{article}
\usepackage[hyperref]{acl2020}
\usepackage{times}
\usepackage{latexsym}
\usepackage[normalem]{ulem}
\usepackage{soul}

\usepackage{microtype}
\usepackage[textsize=tiny]{todonotes}
\aclfinalcopy % Uncomment this line for the final submission
\usepackage{booktabs}
\usepackage{mathrsfs}
\usepackage{makecell}
\usepackage{multirow}
\usepackage{graphicx}
\usepackage{subcaption}
\usepackage{amsthm}
\usepackage{amsmath}
\usepackage{amsfonts}
\usepackage{bm}

\input{math_notations.tex}

\newcommand{\Wei}[1]{\textcolor{black}{#1}}
\newcommand{\metric}[1]{\textsc{#1}}
\newcommand{\se}[1]{\textcolor{black}{#1}}

\title{%Translationese Over Human Reference?? \\
On the Limitations of Cross-lingual Encoders as Exposed by Reference-Free Machine Translation  Evaluation  
}

\author{Wei Zhao$^{\dagger}$, Goran Glava{\v{s}}$^{\ddagger}$,
Maxime Peyrard$^{\Phi}$, Yang Gao$^\star$, Robert West$^{\Phi}$, Steffen Eger$^{\dagger}$\\
    $^\dagger$ 
    Technische Universit\"at Darmstadt %, Germany \\
    $^\ddagger$ 
    University of Mannheim, Germany \\
    $^{\Phi}$
    EPFL, Switerland \
    $^\star$ 
    Royal Holloway University of London, UK \\
    {\tt \{zhao,eger\}@aiphes.tu-darmstadt.de}\\
    {\tt goran@informatik.uni-mannheim.de, yang.gao@rhul.ac.uk} \\
    {\tt \{maxime.peyrard,robert.west\}@epfl.ch}
  }

\begin{document}
\maketitle
\begin{abstract}
\se{Evaluation of cross-lingual %text representations 
encoders 
is usually performed either via zero-shot cross-lingual transfer in supervised downstream tasks or via unsupervised cross-lingual textual similarity.
% . 
%Both are relatively simple setups which require only limited cross-lingual abilities. %, as even bag-of-words models are strong baselines for classification. 
%performs well even with bag-of-words models. 
In this paper, we concern ourselves with 
%with a naturally adversarial setup, namely, 
reference-free machine translation (MT) evaluation  where we directly compare source texts to (sometimes low-quality) system translations, which represents a natural adversarial setup for multilingual encoders.}
Reference-free evaluation %of machine translation 
%This 
holds the promise of web-scale comparison of MT systems. %, across a plethora of genres and domains. 
%Recent WMT evaluations yielded metrics based on cross-lingual semantic representations which exhibited poor correlations with human assessment of translation quality. 
%In this work, 
We systematically investigate a range of metrics based on \se{state-of-the-art} cross-lingual semantic representations obtained with pretrained M-BERT and LASER. %models.   %spaces 
%and 
We find that they perform poorly as semantic encoders for reference-free MT evaluation and
identify their two key limitations, namely,
%most prominently 
(a) a semantic mismatch between representations of mutual translations and, more prominently, (b) the inability to %detect and 
punish ``translationese'', i.e., low-quality %word-by-word 
literal 
translations. 
We propose two partial remedies: 
%for the observed phenomena: 
(1) %a post-hoc weakly-supervised alignment step for improving bilingual semantic alignments in the context of MT evaluation
\se{post-hoc re-alignment of the vector spaces} 
and (2) coupling of semantic-similarity based metrics with target-side language modeling. %, which is inherently suitable for punishing word-by-word translations. 
%We empirically 
%show that the latter leads to 
%demonstrate improvements over established metrics like BLEU,
In segment-level MT evaluation, our best 
% combined 
metric surpasses %the counterparts in the reference-free setup, surpassing 
reference-based BLEU %baseline 
by 5.7 correlation points.
%and 28\% %with human assessment 
%on system-level.}
% \se{On segment-level, our combined metric outperforms BLEU},  
% which operates in the resource-demanding reference-based setting, \se{by 7 points}. 
We make our MT evaluation code available.% whi
\footnote{\url{https://github.com/AIPHES/ACL20-Reference-Free-MT-Evaluation}}
\end{abstract}

\input{1-introduction.tex}
\input{2-related.tex}
\input{3-approach.tex}
\input{4-experiments.tex}
\input{4-analysis.tex}

\input{5-conclusion.tex}
\section*{Acknowledgments}
We thank the anonymous reviewers for their insightful comments and suggestions, which greatly improved the final version of the paper. 
This work has been supported by the German Research Foundation as part of the Research Training
Group Adaptive Preparation of Information from Heterogeneous Sources (AIPHES) at the Technische
Universit\"at Darmstadt under grant No. GRK 1994/1. The contribution of Goran Glava\v{s} is supported by the Eliteprogramm of the Baden-W\"{u}rttemberg-Stiftung, within the scope of the grant AGREE.   

\bibliography{6-references}
\bibliographystyle{acl_natbib}
\clearpage
\input{appendix.tex}
\end{document}

%% file: math_notations.tex
\usepackage{amssymb,amsmath}
\usepackage{mathrsfs}
\newcommand{\R}{\mathbb{R}}

\newcommand{\1}{\boldsymbol{1}}

\newcommand{\bv}{\boldsymbol{v}}

\newcommand{\bx}{\boldsymbol{x}}
\newcommand{\by}{\boldsymbol{y}}
\newcommand{\f}{\boldsymbol{f}}

\def\mC{{\bm{C}}}
\def\mD{{\bm{D}}}

\def\mF{{\bm{F}}}

\def\mQ{{\bm{Q}}}

\def\mU{{\bm{U}}}
\def\mV{{\bm{V}}}
\def\mW{{\bm{W}}}
\def\mX{{\bm{X}}}

\def\ervx{{\textnormal{x}}}
\def\ervy{{\textnormal{y}}}

\newcommand{\WMD}{\text{WMD}}

%% file: 1-introduction.tex
\section{Introduction}

%%%%%%%%%%%%%%%
%GG: Mismatch between the nature of the text generation tasks and the classic fixed-label-set evaluation setup 
%%%%%%%%%%%%%%%

A standard evaluation setup for supervised machine learning (ML) tasks assumes an evaluation metric which compares a %set of model's predictions with a set of gold labels to a classifier prediction.
gold label to a classifier prediction.
%\todo{SE: the previous version said: "a set of labels to a set of predictions" but this is multi-label classification, which is less common, I would say. Or did you mean $E:\{\}\times\{\}\rightarrow \mathbb{R}$? But then $E$ still takes individual instances as input, not sets.}
%\footnote{In settings where collecting gold labels is notoriously difficult, different evaluation paradigms are arguably more suitable---e.g., evaluating classifier's performance by comparing its predictions to predictions of a multitude of other classifiers  \citep{Platanios:2016:EAU:3045390.3045540}}
%\todo{SE: I don't understand this sentence. The standard case is single-label classification, i.e., comparing a classifier prediction to a gold label. Does this talk here about multi-label classification? But even so, I wouldn't understand.} 
This %standard evaluation 
setup %, which dominates the evaluation landscape of natural language processing (NLP) as well, %has one major assumption: 
assumes 
that the task %comes with %a set of 
has clearly defined and unambiguous labels and, in most cases, that an instance can be assigned few labels.
%\todo{SE: you mean multi-label classification or the size of the label set is small?} 
These assumptions, however, do not hold for natural language generation (NLG) tasks like machine translation (MT) \cite{bahdanau2014neural,johnson2017google} and text summarization \cite{rush2015summarization,tan2017abstractive}, where we do not predict a single discrete label but %rather 
generate natural language text. %Since natural language is by definition ambiguous and unconstrained, 
Thus, the set of labels for %text generation tasks 
NLG is neither clearly defined nor finite. Yet, the standard evaluation protocols for NLG %text generation %tasks 
still predominantly follow the described default paradigm: (1) evaluation datasets come with human-created reference texts %(i.e., translations and summaries written by humans) 
and (2) evaluation metrics, e.g., BLEU \cite{Papineni:2002} or METEOR \cite{lavie2007meteor} for MT and ROUGE \cite{lin2003automatic} for summarization, count the exact ``label'' (i.e., $n$-gram) matches between reference %text 
and system-generated text. In other words, established %text generation 
NLG evaluation compares semantically ambiguous %and imprecise 
labels from an unbounded set (i.e., natural language texts) via \se{hard}  symbolic matching (i.e., string overlap).

%%%%%%%%%%%%%%%
%GG: Next we mention the monolingual "semantic" comparison approaches: these remove the "crisp symbolic comparison" problem and allow for semantic comparison, but do not remove the fact that we're comparing the generation with only one of very many possible/correct reference texts. And still require reference text (costly annotation effort) -- reference-free metrics remedy for this.   
%%%%%%%%%%%%%%%
%\todo{SE: in the appendix, the math notation must be fixed in the tables: e.g., $m$ vs. m.}
The first remedy is to replace the \se{hard} symbolic comparison of natural language ``labels'' with %the comparison of texts' meaning, which requires reference texts and system generations to be embedded into the same latent semantic space.
\se{a soft comparison of texts' meaning, using semantic vector space representations.} 
Recently, a number of MT evaluation methods appeared focusing on semantic comparison of reference and system translations \cite{Shimanaka:2018,clark-etal-2019-sentence,zhao2019moverscore}. While these %metrics demonstrate higher correlation 
\se{correlate better} than $n$-gram overlap metrics %like BLEU or METEOR 
with human assessments, 
%of semantic alignment between reference and system translations, 
they do not address %the issues inherent to the need for having reference translations, 
inherent limitations stemming from the need for reference translations,
namely: %reference translations 
(1) references %translations 
are expensive to obtain; (2) %a reference translation assumes 
they 
assume 
a single correct solution and bias %es 
the evaluation, both automatic and human \cite{dreyer2012hyter,fomicheva2016reference}, 
%\todo{SE: I think in summarization, people use multiple references. SE-2:  yes, but with DA=CLDA, we could not confirm this bias in our experiments, right} 
and %that 
(3) limitation of MT 
evaluation %only 
to %the 
language pairs with available parallel data.       
%\todo{SE:?}

%%%%%%%%%%%%%%%
%GG: Now motivating reference-free MT evaluation metrics    
%%%%%%%%%%%%%%%
Reliable \emph{reference-free} evaluation metrics, directly measuring the (semantic) correspondence between the 
source language 
text and %the 
system translation, %without the need for references, 
would  remove the need for human references %translations 
%in MT evaluation 
and 
allow for unlimited MT evaluations: 
%\Wei{very widely applicable} in scope: 
%besides for training MT models, 
\emph{any} %parallel 
monolingual 
corpus could %also 
%directly 
be used for evaluating %the 
MT systems.
%\todo{SE: everything after the colon, I don't understand - with reference-free evaluation, we free ourselves from the need of (human-generated) parallel corpora. This means we can evaluate on everything, we only need monolingual data.} 
However, the proposals of reference-free MT evaluation metrics have been few and far apart
%\todo{SE: in time?: GG "few and far apart" is an idiomatic expression} 
and have required either non-negligible supervision (i.e., human translation quality labels) \cite{specia2010machine} or language-specific preprocessing like semantic parsing \cite{lo-2014-xmeant,lo-2019-yisi}, both hindering the wide applicability of the proposed metrics. Moreover, they have also typically exhibited performance levels well below those of standard reference-based metrics \cite{ma-etal-2019-results}.
%limiting their large-scale usability. 

%%%%%%%%%%%%%%%
%GG: Now our contributions
%%%%%%%%%%%%%%%
In this work, we comparatively evaluate a number of reference-free MT evaluation metrics that %rely 
build on the most recent developments in multilingual representation learning, namely cross-lingual %(static and contextualized) 
\se{contextualized}  
embeddings %\cite{conneau2018word,Devlin:2018} 
\cite{Devlin:2018} 
and cross-lingual sentence encoders \cite{artetxe2019massively}. %,yang2019multilingual}. %and require no supervision nor any language-specific preprocressing.\todo{SE: we require no supervision? Except when we post-align MBERT?} 
We investigate two types of cross-lingual reference-free metrics: \textbf{(1)} \textit{Soft token-level alignment} metrics find the optimal soft alignment between source sentence and system translation %token 
using %the 
Word Mover's Distance (WMD) \cite{kusner2015word}. \newcite{zhao2019moverscore} recently demonstrated that WMD operating on %contextualized  
BERT representations \cite{Devlin:2018}
%-contextualized \todo{SE: BERT-contextualized?} 
substantially outperforms baseline MT evaluation metrics in the %monolingual 
reference-based setting. In this work, we investigate whether WMD can yield comparable success
%\st{useful} 
in the %cross-lingual (i.e., reference-free) 
reference-free (i.e., cross-lingual) setup;
%, applied on top of cross-lingual token representations. 
\textbf{(2)} \textit{Sentence-level similarity} metrics measure the similarity between sentence representations of the source sentence and system translation \se{using cosine similarity}.  
%, as produced by multilingual sentence encoders like LASER \cite{artetxe2019massively} and USE \cite{cer-etal-2018-universal,yang2019multilingual}. 

Our analysis yields several interesting findings. 
\textbf{(i)} We show that, unlike in the monolingual reference-based setup, %\cite{zhao2019moverscore}, 
%the cross-lingual reference-free evaluation metrics fail to outperform the reference-based baseline metrics like BLEU.
metrics that operate on %static or 
contextualized %text 
representations generally do not outperform symbolic matching metrics like BLEU, which operate in the reference-based environment. 
%\todo{YG: even in monolingual setup, reference-free metrics underperform reference-based ones, don't they? The MoverScore paper does not study any reference-free metrics.}
\textbf{(ii)} We identify two %possible 
reasons for this failure: 
(a) firstly, 
cross-lingual semantic mismatch, %a mismatch of cross-lingual representations, 
especially for multi-lingual BERT (M-BERT), which 
%builds cross-lingual
construes a shared multilingual space in an unsupervised fashion, without any direct bilingual signal; 
%(iii)  
%We next show that there is still a significant
(b) secondly, the inability of the %both above classes of 
\se{state-of-the-art} cross-lingual metrics based on multilingual encoders to adequately capture and 
punish ``translationese'', i.e., literal word-by-word translations of the source sentence---as translationese is an especially persistent property of MT systems, this problem is particularly troubling in our context of reference-free MT evaluation. 
%\todo{@Wei: can you especially check whether this claim is true?: GG: wasn't true, fixed it now.} 
%and which leads to a word-alignment mismatch. 
%language bias in the cross-lingual representations produced by state-of-the semantic encoders like Multilingual BERT \cite{devlin2019bert} and LASER \cite{artetxe2019massively} (vectors of mutual translations are insufficiently close) and that 
\textbf{(iii)} We show that by executing an additional weakly-supervised cross-lingual %alignment step, we 
re-mapping step, we can to some extent alleviate both previous issues. 
%can improve the performance of all reference-free metrics under investigation. 
\textbf{(iv)}  
Finally, we show that the combination of cross-lingual reference-free metrics and language modeling on the target side (which is able to detect ``translationese''), %\todo{SE: I agree that the LM detects translationese. In the paper, however, we pretend that the re-mapping is the big star.} 
surpasses the performance of reference-based baselines.

%%% GG: Well done Steffen for the below paragraph!
%
\se{Beyond designating a viable prospect of web-scale domain-agnostic MT evaluation, our findings indicate that the challenging task of reference-free MT evaluation is able to expose an important limitation of current state-of-the-art multilingual encoders, i.e., the failure to properly represent corrupt input, that may go unnoticed in simpler evaluation setups such as zero-shot cross-lingual text classification or measuring cross-lingual text similarity not involving %such 
``adversarial'' conditions.}  \Wei{We believe this is a promising direction %to facilitate 
for nuanced, fine-grained evaluation of  
cross-lingual representations, %together with 
extending the recent benchmarks which focus on zero-shot transfer scenarios \cite{Hu:abs-2003-11080}.}

%% file: 2-related.tex
\section{Related Work}
\label{sec:related}
Manual human evaluations of MT systems undoubtedly yield the most reliable results, but are expensive, tedious, and generally do not scale to a multitude of domains. A significant body of research is thus dedicated to the study of automatic evaluation metrics for machine translation. Here, we provide an overview of both reference-based MT evaluation metrics and 
% present %some important ones and 
recent research efforts towards reference-free MT evaluation, which leverage cross-lingual semantic representations and unsupervised MT techniques.

%\todo{SE: reference-free or reference-less?} 
%\todo[inline]{I think it would be justified to talk about unsupervised MT also - Artetxe et al. MP: Yes, I added some references, maybe we can put more?}

\paragraph{Reference-based MT evaluation.}
Most of the commonly used evaluation metrics in MT compare system and reference translations. They are often based on surface forms such as $n$-gram overlaps like BLEU \cite{Papineni:2002}, SentBLEU, NIST \cite{Doddington:2002}, chrF++ \cite{popovic-2017-chrf} or METEOR++\cite{guo-hu-2019-meteor}. They have been extensively tested and compared in recent WMT metrics shared tasks \cite{Bojar:2017,Ma:2018, ma-etal-2019-results}.

These metrics, however, operate at the surface level, and by design fail to recognize semantic equivalence lacking lexical overlap. To overcome these limitations, some research efforts exploited static word embeddings \cite{MikolovSCCD13} and trained embedding-based supervised metrics on sufficiently large datasets with available human judgments of translation quality \cite{Shimanaka:2018}. 
With the development of contextual word embeddings \cite{Peters:2018,Devlin:2018}, we have witnessed proposals of semantic metrics that account for word order. For example, \newcite{clark-etal-2019-sentence} introduce a semantic metric relying on sentence mover’s similarity and the contextualized ELMo embeddings \cite{Peters:2018}. 
Similarly, \newcite{zhang:2019} describe a reference-based semantic similarity metric based on contextualized BERT representations \cite{Devlin:2018}. \newcite{zhao2019moverscore} generalize this line of work with their MoverScore metric, which computes the mover's distance, i.e., the optimal soft alignment between tokens of the two sentences, based on the similarities between their contextualized embeddings. \Wei{\newcite{mathur-etal-2019-putting}} %investigate a supervised metric for reference-based MT evaluation, which is a BERT-based regressor trained on WMT human evaluation datasets.}
\se{train a supervised BERT-based regressor for reference-based MT evaluation.}
\paragraph{Reference-free MT evaluation.} 
Recently, there has been a growing interest in reference-free MT evaluation \cite{ma-etal-2019-results}, \se{also referred to as ``quality estimation'' (QE) in the MT community}. In this setup, evaluation metrics semantically compare system translations directly to the source sentences. The attractiveness of automatic reference-free MT evaluation is obvious: it does not require any human effort or %any 
parallel data.   
%beneficial because obtaining human-written references is costly and time-consuming. 
%\todo[inline]{SE: we should add 1 or 2 references from ealier research, e.g., XMeant. MP: Yes, added}
%
To approach this task, \Wei{\newcite{popovic-etal-2011-evaluation} exploit a bag-of-word translation model to estimate translation quality, which sums over the likelihoods of aligned word-pairs between source and translation texts.
\newcite{specia-etal-2013-quest} estimate translation quality using language-agnostic linguistic features extracted from source lanuage texts and system translations.}
\newcite{lo-2014-xmeant} introduce XMEANT as a cross-lingual reference-free variant of MEANT, a metric based on semantic frames. \newcite{lo-2019-yisi} extended this idea by leveraging M-BERT embeddings. \Wei{The resulting metric, YiSi-2, evaluates system translations by summing %the 
similarity scores over words pairs that are best-aligned mutual translations. 
% In contrast, our  proposed metric softens this greedy alignment and allows for matching a small amount of source language words to semantically corresponding words in system translations by solving a constrained optimization problem.
% via word-level semantic similarity.  %While 
YiSi-2-SRL optionally
% A modification proposed in the same paper, 
% YiSi-2-srl, 
combines an additional similarity score based on the  alignment over the semantic structures (e.g., semantic roles and frames).
% of source language texts and system translations.
% like semantic roles and frames.
% \todo{SE: the difference between Yisi and our approach is the lack of remapping and the absence of the language model} 
Both metrics are %indeed
reference-free, but YiSi-2-SRL is not %generally 
resource-lean as it requires %the availability of the 
a semantic parser for both languages.
%Moreover, 
Moreover, in contrast to \Wei{our proposed metrics}, %our proposed metric 
they do not mitigate the misalignment of cross-lingual embedding spaces and %combines perplexity scores yielded by a LM against translationese.
do not integrate a target-side language model, \se{which we identify to be crucial components}. 
}
% \todo{SE: what about YiSi-2? What is the main difference to our work?}
%\todo{SE: if this is also unsupervised, what's the difference to us? WMD?}
%Furthermore, if such metrics correlate highly with humans, it can directly be optimized as the objective function of a translation system. Thus, ensuring strong agreement between the system's output and human judgments.

Recent progress in cross-lingual semantic similarity \cite{agirre-etal-2016-semeval,cer-etal-2017-semeval} and unsupervised MT \cite{artetxe2019massively} 
%can be helpful.
has also led to novel reference-free metrics. 
For instance, \newcite{yankovskaya-etal-2019-quality} propose to train a metric combining multilingual embeddings extracted from M-BERT and LASER \cite{artetxe2019massively} together with the log-probability scores from neural machine translation.
%\todo{SE: it would be important to know how we differ from those two works} 
Our work differs from  that of \newcite{yankovskaya-etal-2019-quality} in one crucial aspect: the cross-lingual reference-free metrics that we investigate and benchmark do not require any human supervision.
% Our work differs from the metrics of \newcite{yankovskaya-etal-2019-quality} in two important aspects: (1) the cross-lingual reference-free metrics we investigate and benchmarks are fully unsupervised and (2) \Wei{our method does not require} computationally expensive training of unsupervised MT systems.\todo{SE: who is training unsupervised MT systems?}
%
%\Wei{
%\paragraph{Cross-lingual Representations} Pretraining cross-lingual representations over a number of languages has the potential to encode these languages into a shared embedding space. For instance, pretraining M-BERT \cite{Devlin:2018} involves the concatenation of monolingual data in diverse languages while LASER \cite{artetxe2019massively} is jointly pretrained on parallel corpora of 93 languages. However, in lower resourcing languages, pretraining M-BERT with little pretraining data yields poorer quality representations \cite{Conneau:2019a}. \citet{Cao:2020} recently show the mutual translations of word pairs are considerably misaligned in the embedding space. Our findings in addition show that cross-lingual embeddings favor translationese over human refehttps://www.overleaf.com/project/5dc30c8e63726500011e82e5rences. All these reveal the limitations of cross-lingual representations.}
\paragraph{Cross-lingual Representations.} 
Cross-lingual text representations offer a prospect of modeling meaning across languages and support cross-lingual transfer for downstream tasks \citep{klementiev-etal-2012-inducing,rueckle:2018,glavas-etal-2019-properly,josifoski2019crosslingual, Conneau:2020acl}.  
Most recently, the (massively) multilingual encoders, such as multilingual M-BERT \cite{Devlin:2018}, XLM-on-RoBERTa \cite{Conneau:2020acl}, and (sentence-based) LASER, have  profiled themselves as state-of-the-art solutions for (massively) multilingual semantic encoding of text. %Both induces cross-lingual spaces involving about 100 languages.
While LASER has been jointly trained on parallel data of 93 languages, M-BERT has been trained on the concatenation of monolingual data in more than 100 languages, without any cross-lingual mapping signal. There has been a recent vivid discussion on the cross-lingual abilities of M-BERT \citep{pires-etal-2019-multilingual,K2020Cross-Lingual,Cao2020Multilingual}. In particular, \citet{Cao2020Multilingual} show that M-BERT often yields disparate vector space representations for mutual translations and propose a multilingual re-mapping based on parallel corpora, to remedy for this issue. In this work, we introduce re-mapping solutions that are resource-leaner and require easy-to-obtain limited-size word translation dictionaries rather than large parallel corpora.

%% file: 3-approach.tex
\section{Reference-Free MT Evaluation Metrics }
%In this work, we %abstract away 
%free ourselves 
%from the need for costly human references in MT evaluation by introducing a ``reference-free'' metric. We now describe our metric in detail.
% named XMoverScore that can be seen as a variant of MoverScore~\cite{DBLP:conf/emnlp/no_reference19}.

% cross-lingual variant of MoverScore.
% using cross-lingual metric.
% replacing the monolingual word embeddings in MoverScore with cross-lingual embeddings~\cite{lample2019cross,Artetxe:2018}. 

% investigate three approaches to compare neural and non-neural MT systems in reference-free condition: 

% Let $\bx$ and $\by$ be two sentences viewed as sequences of $n$-grams: $\bx^n$ and $\by^n$.
% be source sentence viewed as a sequence of words. We denote by $\bx^n$ the sequence %list 
% of $n$-grams of $\bx$ (i.e., $\bx^1=\bx$ is the sequence %list 
% of words and $\bx^2$ is the sequence of bigrams).
%\subsection{Notation}
In the following, we use $\mathbf{x}$ to %refer to 
denote a source sentence (i.e., a sequence of tokens in 
the source language),
%tokens) 
$\mathbf{y}$ to denote a system translation of $\mathbf{x}$ in the 
target language, 
and $\mathbf{y}^\star$ to denote the human
reference translation for $\mathbf{x}$. 
%Thus, we have triples $(\mathbf{x},\mathbf{y},\mathbf{y}^\star)$. 
%YG: why we have y^star? aren't we study ref-free metric? if use DA use ground-truth, we do not need y^star, right?

\subsection{Soft Token-Level Alignment} 
We start from the MoverScore~\cite{zhao2019moverscore}, a recently 
%introduced 
proposed reference-based MT evaluation metric designed to measure 
the semantic similarity %distance 
between system outputs ($\mathbf{y}$) %translation 
and human references ($\mathbf{y}^\star$). 
It finds an optimal soft semantic alignments between tokens from $\mathbf{y}$ and 
$\mathbf{y}^\star$ by minimizing the 
Word Mover's Distance \cite{kusner2015word}. 
% \todo{YG: @Wei, check whether this is correct.}
%It implicitly aligns semantically similar words %and 
%by %finding the amount of flow traveling between them.\todo{SE: flow is too general} %these words. 
%determining the optimal transformation between two sets in terms of `transportation costs'. 
%Inspired by the success of MoverScore, 
%
In this work, we extend the MoverScore metric to operate in the cross-lingual
setup, i.e., to measure the semantic similarity between
$n$-grams (unigram or bigrams) of the source text $\mathbf{x}$ and the  system translation $\mathbf{y}$, represented with embeddings originating from a cross-lingual semantic space.
%
%to compare semantic similarity between system translation and source text. 
%for comparing system output $\mathbf{y}$ to source text $\mathbf{x}$. 
%In the following, we formulate Cross-lingual MoverScore operating on $n$-grams.
% \todo{SE: let's use cross-lingual rather than multi-lingual}
% \todo{SE: some package appears to be missing here. See error messages.}

%Let $\bx$ and $\by$ be source sentence and system translation viewed as sequences of $n$-grams: $\bx^n$ and $\by^n$, i.e., $\bx_n = (\ervx_1^{n}, \dots, \ervx_m^{n})$. 
First,  we decompose the source text $\mathbf{x}$ into a sequence of $n$-grams, 
denoted by $\bx_n = (\ervx_1^{n}, \dots, \ervx_m^{n})$ and then do the same operation for the system translation $\mathbf{y}$, denoting the resulting sequence of n-grams with $\by_n$.
%
%then we define a distance metrix $C$ such that 
%If we have a Euclidean distance metric between $n$-grams,
% \todo{SE: $d$ has disappeared. $d=||\cdot||_2$?} 
Given $\bx_n$ and $\by_n$,
we can then define a distance matrix $\bm{C}$ such that 
$\bm{C}_{ij} =  \|E(\ervx^n_i) - E(\ervy^n_j) \|_2$ 
is the distance between the $i$-th $n$-gram of $\bx$ 
and the $j$-th $n$-gram of $\by$, 
where $E$ is a \emph{cross-lingual} embedding function that
maps %$n$-grams 
\se{text} 
in %disparate 
different languages to %their vector representations in 
a shared embedding space.
% \todo{SE: there are several instances below where $E$ also maps sentences so I rewrote to "text". Additionally, I wonder how bigrams are mapped to a vector?} 
%
With respect to the function $E$, we %comparatively 
experimented with cross-lingual representations induced (a) from static word embeddings with RCSLS \cite{joulin-etal-2018-loss}) 
(b) with M-BERT~\cite{Devlin:2018} as the multilingual encoder; with a focus on the latter. For M-BERT, we take the representations of the last transformer layer as the %cross-lingual %token 
text 
representations. 

$\WMD$ between the two sequences of $n$-grams $\bx^n$ and $\by^n$ with associated $n$-gram weights \footnote{We follow %the operations %proposed in MoverScore 
\citet{zhao2019moverscore} in obtaining  $n$-gram embeddings and their associated weights based on IDF.} to  $\f_{\bx^n}\in\R^{|\bx^n|}$ and $\f_{\by^n}\in\R^{|\by^n|}$ is defined as:
% \todo{MP: At this point, the reader may wonder about what are the n-gram weights and were they come from.}
\begin{align*}
& m(\bx,\by):= \WMD(\bx^n,\by^n)
= \min_{\mF} \sum_{ij} \mC_{ij}\cdot \mF_{ij}, \\
&\text{s.t. } \mF\1=\f_{\bx^n},\;\;\mF^{\intercal}\1=\f_{\by^n},
\end{align*}
where $\mF \in\R^{|\bx^n|\times |\by^n|}$ is a transportation matrix with $\mF_{ij}$ denoting the amount of flow traveling from 
%the $i$-th $n$-gram %YG: these too many hyphens do not help but instead confuse the readers;  
$\ervx^n_i$ to %in $\bx^n$ to the $j$-th $n$-gram 
$\ervy^n_j$.
% \todo{SE: which representations are you taking?}
%in $\by^n$. 

\subsection{Sentence-Level Semantic Similarity}

In addition to measuring semantic distance between %system translation and human reference 
$\mathbf{x}$ and $\mathbf{y}$ 
at word-level, one can also encode them into sentence representations with multilingual sentence encoders like LASER~\cite{artetxe2019massively}, and then measure their cosine distance
%compute the distance between 
%two sentences in %disparate 
%distinct languages 
%with cosine similarity: %, denoted as:
\begin{align*}
m(\bx,\by) = 1 - \frac{E(\bx)^{\intercal} E(\by)}{ \|E(\bx) \| \cdot  \|E(\by) \|}.
\end{align*}
%\se{where $E(\mathbf{x})$ now yields the sentence vector for $\mathbf{x}$}. 
% \todo{YG: @Wei: NOTICE! you want distance, so should use cosine distance instead of cosine similarity, right? Hence I add the '1-' part}

\subsection{Improving Cross-Lingual Alignments}
\label{sec:cla}

% \begin{figure*}
% \centering
% \includegraphics[width=\linewidth]{figures/alignment2.pdf}
% \caption{\todo[inline]{SE: I suggest to remove this picture, as I do not believe the explanations are correct} PCA Visualization of sentence embeddings of source sentence in German and reference translation in English. In the initial sentence embedding space encoded from FastText,\todo{SE: FastText hasn't been discussed before. Before, you said the alignment is bad, but now the spaces are independent, i.e., there is no mapping} 
% source and reference embeddings are not aligned. While they are seemingly aligned in the cross-lingual space encoded from LASER with an implicit bilingual dictionary populated by shared words/wordpieces across languages (middle), these embeddings are ill-aligned when zooming into individual instances (right), where source sentences  are often close to translationese (word-by-word translation) instead of reference translation. %, which is less preferred by human raters. 
% Ideally, with post-hoc cross-lingual alignment, one can mitigate this misalignment by projecting source sentence into reference translation.}
%  \label{fig:alignment_visualization}
% \vspace{-0.15in}
% \end{figure*} 
%resulting 

%Our 
Initial analysis indicated that, despite the multilingual pretraining of M-BERT \cite{Devlin:2018} and LASER \cite{artetxe2019massively}, the monolingual subspaces of the multilingual spaces they induce are far from being semantically well-aligned, i.e., we obtain fairly distant vectors for mutual word or sentence translations.\footnote{
LASER is jointly trained on parallel corpora of different languages, but in resource-lean language pairs,
the induced embeddings from mutual translations may be far apart. %for language pairs with little parallel data.
% that are typologically dissimilar.
% and have a considerate difference in the sizes of pretraining corpora.
} %; \se{see also \citet{schuster-etal-2019-cross}}. 
To this end, we apply two simple, weakly-supervised linear projection methods for post-hoc improvement of the cross-lingual alignments in these multilingual representation spaces.

\paragraph{Notation.}
Let $\mD=\{(w_{\ell}^1,w_{k}^1), \dots, (w_{\ell}^n,w_{k}^n)\}$ be a set of 
%bilingual dictionary with either word or sentence pairs provided or unsupervisedly learned from $\ell_i$th and $\ell_j$th language corpus. 
matched word or sentence pairs from two different languages $\ell$ and $k$. 
We define a %debiasing 
re-mapping
function $f$ such that any $f(E(w_{\ell}))$ and $E(w_{k})$ are better aligned in the resulting shared vector space. We investigate two resource-lean choices for the re-mapping function $f$.

% For instance, as proposed by Aldarmaki & Diab (2019), one
% method to perform alignment on contextualized representations is to first use word alignment pairs
% extracted from parallel corpora as a dictionary, learn an alignment matrix W based on it, and apply
% W back to the extracted representations.

% \paragraph{Generalized Bias-Direction Debiasing (GBDD)}
\paragraph{Linear Cross-lingual Projection (CLP).}
\se{Following related work \cite{schuster-etal-2019-cross}, we re-map contextualized embedding spaces using linear projection.}  
Given $\ell$ and $k$, %one can 
we stack all vectors of the source language words and target language words for pairs $\mD$, respectively, to form matrices $\mX_{\ell}$ and $\mX_{k} \in \mathbb{R}^{n\times d}$, with $d$ as the embedding dimension and $n$ as the number of word or sentence alignments. 
The word pairs we use to calibrate M-BERT 
are extracted from 
EuroParl \cite{koehn2005europarl} using FastAlign \cite{dyer-etal-2013-simple},
and the sentence pairs to calibrate LASER are sampled
directly from EuroParl.\footnote{While LASER requires large parallel corpora in pretraining, 
we believe that 
fine-tuning/calibrating the embeddings post-hoc requires
fewer data points.
%we assume that re-mapping instead needs a rather small amount.
}
%which has the same goal of remedying cross-lingual alignment in recent work \cite{aldarmaki-diab-2019-context, schuster-etal-2019-cross, cross-lingual-wang:2019}.} 
%For M-BERT, we take the representations of the last transformer layer as the cross-lingual token representations.
\citet{mikolov13} propose to learn a projection matrix $W \in \mathbb{R}^{d\times d}$ 
% \se{where $d$ is the embedding size,}\todo{SE: $n$ is the size of the vocabulary? Note that this is the same transformation as done in Schuster et al. - however, it is unclear to me how your mapping proceeds for MBERT - which representation are you using; there are 12 layers? Do you count the same words in different contexts as different instances?} 
% such that $\mW \mX_{\ell_1}$ can approximate $\mX_{\ell_2}$, 
%with SGD optimizer,
by minimizing the Euclidean distance beetween the projected source language vectors and their corresponding target language vectors:
%denoted as:
% To find the optimal projection matrix $W \in \mathbb{R}^{d\times d}$,  proposed to solve the following optimization problem:
\begin{equation}
    \min_{\mW}  \| \mW \mX_{\ell} - \mX_{k} \|_2. \nonumber
\end{equation}
\citet{xing-etal-2015-normalized} %latter 
achieve further improvement on the task of bilingual lexicon induction (BLI) by constraining $\mW$ to an orthogonal matrix, i.e., such that $\mW^\intercal \mW=\mathbf{I}$. This turns the
% \todo{SE: W needs to be bold-font} 
optimization into the well-known Procrustes problem \cite{schonemann1966generalized} 
% \todo{SE: what is the Procustes problem? Wei: added citation} 
with the following closed-form solution:
\begin{align} \label{eq:procrustes}
    \hat{\mW} & = \mU\mV^\intercal, 
    \mU\Sigma \mV^\intercal = \text{SVD}(\mX_{\ell}\mX_{k}^\intercal) \nonumber
\end{align}
We note that the above CLP re-mapping is known to have deficits, i.e., it requires the embedding spaces of the involved languages to be approximately isomorphic \citep{sogaard-etal-2018-limitations,vulic2019we}. Recently, some re-mapping methods that reportedly remedy for this issue have been suggested \cite{glavavs2020instamap,mohiuddin2020lnmap}. We leave the investigation of these novel techniques for our future work.
\iffalse
\Wei{
The post-hoc re-mapping is based on a strong assumption that the embedding spaces of the
considered languages are approximately \emph{isomorphic}, i.e.,\ 
sharing the same underlying structure ~\citep{sogaard-etal-2018-limitations}, 
which may not hold for cross-lingual contextualized embeddings. %because... 
\citet{Cao:2020} replace the post-hoc 
re-mapping with direct fine-tuning 
on M-BERT by enforcing selected words
to align, which potentially allows 
for learning alignment of hundreds 
of languages simultaneously. 
%
In depth comparison of post-hoc 
re-mapping and fine-tuning is beyond
the scope of this paper, and we 
leave it for future work.
%We leave the further study of these matters in future work.
}\todo{SE: this needs to be rewritten Wei: I am not confident to write this para. YG: make some changes. Do give a go to make sure the opinions are correct}
\fi

% that embedding spaces are approximately isometric~\citep{sogaard-etal-2018-limitations}, which may not hold for multilingual contextual representations, it still allows for aligning any language pairs by involving a pivot language, i.e., remapping source-pivot and pivot-target.}
% where $\hat{W}$ denotes the optimal solution and SVD($\cdot$) stands for the Singular Value Decomposition. Finally, the re-mapping function is the linear projection given by $\hat{W}$. \se{This re-mapping approach is identical to that of \citet{schuster-etal-2019-cross}.}
%In this approach, the re-mapping function $f$ is a linear projection $\hat{W}$.

\paragraph{Universal Language Mismatch-Direction (UMD)}
%\todo{SE: is this the official name? It sounds like incorrect English...}
Our second post-hoc linear alignment method is inspired by the recent work on removing biases in distributional word vectors \citep{dev-phillips-2019,lauscher2019general}. We adopt the same approaches in order to quantify and remedy for the ``language bias'', i.e., representation mismatches between mutual translations in the initial multilingual space. 
% bias
%direction in semantics between representations of source and human references .
% the misalignment between the representations of 
% most informative bias direction such that 
% accounting for the misalignment in the cross-lingual embedding space.\todo{SE: again, alignment/misalignment is used too frequently here, and it is not well-specified}
%
Formally, given
$\ell$ and $k$, we create individual misalignment vectors  $E(w_{\ell}^i)-E(w_{k}^i)$ for each bilingual pair in $\mD$. Then we stack these individual vectors to form a matrix $\mQ\in \mathbb{R}^{n\times d}$. 
We then obtain the global misalignment vector $\bv_B$ as the top left singular vector of $\mQ$. The global misalignment vector presumably captures the direction of the representational misalignment between the languages better than the individual (noisy) misalignment vectors $E(w_{\ell}^i)-E(w_{k}^i)$. %  most semantically mismatched direction.
% \todo{SE: why has bias-direction a hyphen?}
%\todo{SE: a reference is needed. What is the dimensionality of Q? Wei: added}
%
Finally, we modify all vectors $E(w_{\ell})$ and $E(w_k)$, by subtracting %from them 
their projections onto the global misalignment direction vector $v_B$:
\begin{equation}
f(E(w_{\ell})) = E(w_{\ell}) - \text{cos}(E(w_{\ell}), v_B)v_B. \nonumber
\end{equation}

\paragraph{Language Model} 
%After re-evaluating 
%reference-free metrics, 
% \Wei{\citet{gamon2005sentence} 
% show a strong correlation between
% language model's perplexity scores and 
% human ratings on translation quality.
% %investigate how correlated language model perplexity scores are with human assessments of translation quality.
% }
BLEU scores
% on the other hand, 
often
fail to reflect the fluency level of translated
texts \cite{edunov-2019}.
%provide evidence that BLEU often fails to %assessing 
%capture the improved fluency of back-translations;
%they thus propose to utilize a language model as 
%the fluency scorer that would complement BLEU.
Hence, we %adopt this idea and 
use the language model (LM) of the 
target language to regularize 
the cross-lingual semantic similarity metrics,
% , which can punish translationese. 
%in reference-free evaluation, since reference-free metric potentially struggle in assessing the fluency of word-by-word and code-switching translation, 
by coupling our cross-lingual similarity scores with a GPT language model of the target language \cite{radford2018improving}. 
%as a fluency scorer. 
We expect the language model to penalize translationese, i.e., unnatural
word-by-word 
% and code-switching 
translations and boost the performance of our metrics.\footnote{We linearly combine the cross-lingual metrics with the LM scores using a coefficient of 0.1 for all setups.
% the LM, for all languages, WMT datasets and cross-lingual embeddings. 
We choose this value based on initial experiments on one language pair.}

%% file: 4-experiments.tex
\section{Experiments}
In this section, we evaluate the quality of our MT reference-free metrics by correlating 
% \todo{SE: I know the word "curate" only in the context of datasets...} 
them with human judgments of translation quality. These quality judgments are based on comparing human references and system predictions. %see also our discussion 
We will discuss this discrepancy 
in \S\ref{sec:human}.  %The cross-lingual similarity metrics we evaluate are as follows: 
%We experiment %our metric 
%along two dimensions: 
% \todo{MP: what are the three dimensions here? We describe only two levels here: Word and Sentence}
% in the following:
% In the experiment, we investigate reference-free metrics along two dimensions: (i) the granularity of embeddings, e.g., static and contextualized word embedding as well as sentence embedding (ii) the granularity of cross-lingual alignment. 
\paragraph{Word-level metrics.} We denote our word-level alignment metrics based on WMD \se{as}  \metric{MoverScore-ngram + Align(Embedding)}, where \metric{Align} is one of our two post-hoc cross-lingual alignment methods (CLP or UMD). For example, \metric{Mover-2 + UMD(M-BERT)} denotes the metric combining MoverScore based on bigram alignments, %exemplifying 
\se{with} 
M-BERT embeddings and UMD as the post-hoc alignment method. 
%and aligned multilingual BERT to measure the semantic distance between two sets of bigram-based word embeddings. 

\paragraph{Sentence-level metric.} We denote our sentence-level metrics \se{as}: %notation containing two ingredients as: 
\metric{Cosine + Align(Embedding)}. For example, \metric{Cosine + CLP(LASER)} measures the cosine distance between the sentence embeddings obtained with LASER, post-hoc aligned with CLP.

% \paragraph{Cross-lingual Alignment} We denote our reference-free metric with two pas:
% %notation containing three ingredients as: 
% % with three ingredients as:\todo{SE: ingredients is not a good word}
% Metric + Alignment(Embedding). For example, Cosine + CLP(LASER) is to measure the cosine similarity between two cross-lingual aligned LASER sentence embeddings
% , where the source sentence embedding is debiased using UMD followed by CLP.

% XMoverScore + FastText
% LASERScore 
% USEScore
% evaluate the quality of MT reference-free metrics by studying the correlations  between these metrics and human judgments.
% We employ two text encoders to embed $n$-grams:
% $\bertbase$, which uses a 12-layer transformer,
% and $\elmobase$, which uses a 3-layer BiLSTM.
\subsection{Datasets}
We collect the source language sentences, their system and reference translations from the WMT17-19 news translation shared task \cite{bojar-etal-2017-results,ma-etal-2018-results,ma-etal-2019-results}, which contains predictions of 166 translation systems across 16 language pairs in WMT17, 149 translation systems across 14 language pairs in WMT18 and 233 translation systems across 18 language pairs in WMT19. We evaluate for X-en language pairs, selecting X from a set of 12 diverse languages: German (de), Chinese (zh), Czech (cs), Latvian (lv), Finnish (fi), Russian (ru), and Turkish (tr), Gujarati (gu), Kazakh (kk), Lithuanian (lt) and Estonian (et). Each language pair in WMT17-19 has approximately 3,000 source sentences, each associated to one reference translation and to the automatic translations generated by participating systems.
% \todo{SE: Wasn't it WMT2019 Wei: The results in WMT2019 are not ready..}

\subsection{Baselines}
We compare with a range of reference-free metrics: ibm1-morpheme and ibm1-pos4gram \cite{popovic-2012-morpheme}, LASIM \cite{yankovskaya-etal-2019-quality}, LP \cite{yankovskaya-etal-2019-quality}, YiSi-2 and YiSi-2-srl \cite{lo-2019-yisi}, and reference-based baselines BLEU \cite{Papineni:2002}, SentBLEU \cite{koehn-etal-2007-moses} and ChrF++ \cite{popovic-2017-chrf} for MT evaluation (see \S\ref{sec:related}).\Wei{\footnote{The code of these unsupervised metrics is not released, thus we %have to 
compare to their official results on WMT19 only.}}
% These metrics were officially reported on the WMT19 shared task \cite{ma-etal-2019-results}. For several of them we could not find available code. Thus, we compare with their performances reported in the WMT2019 benchmark.
%alone using the 
%For the metrics where no code was released, we copy the performance reported by the WMT2019 shared task.
%For the above reference-free baselines reported by the WMT2019 shared-task, we cannot find the released code. 
%Thus, we compare with them in the WMT2019 benchmark alone.
% for the above reference-free baselines reported by the WMT2019 shared-task, code is not published. Thus, we compare 
% participating in WMT2019 
%we could not find code, 
% % code is seemingly not published.\todo{SE: it is important to know the differences between these metrics and ours}
% Thus, we %, therefore, 
% compare our metrics with numbers reported by the WMT2019 shared-task \cite{ma-etal-2019-results}.
% only compare reference-free baselines for WMT2019 because we do not find released code. 
The main results are reported on WMT17. We report the results obtained on WMT18 and WMT19 in the Appendix.
% Due to space constraints, we report the main results on WMT17
% % segment- and system-level 
% and system-level results on WMT19. The results on WMT18 and segment-level results on WMT19 are reported in the appendix.
% only compare with the strongest baselines, the rest can be found in the appendix.

\input{tables/wmt17-segment}
\subsection{Results}
\begin{figure}
\centering
\includegraphics[width=0.85\linewidth]{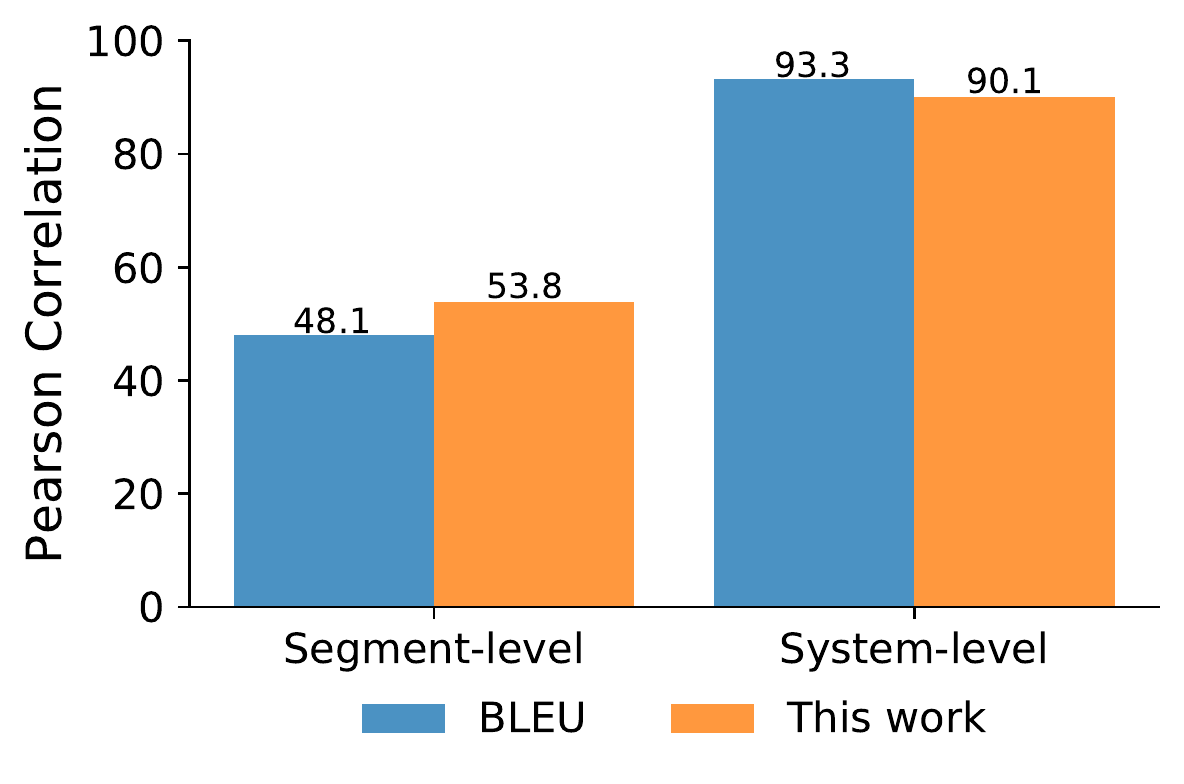}
\caption{Average results of our best-performing metric, together with reference-based BLEU on WMT17. 
% Results are averaged over seven language pairs.
}
% \se{SE: write segment level, not Seg-Level}}
\label{fig:wmt_17_sys_seg}
\end{figure}

\begin{figure}

\includegraphics[width=\linewidth]{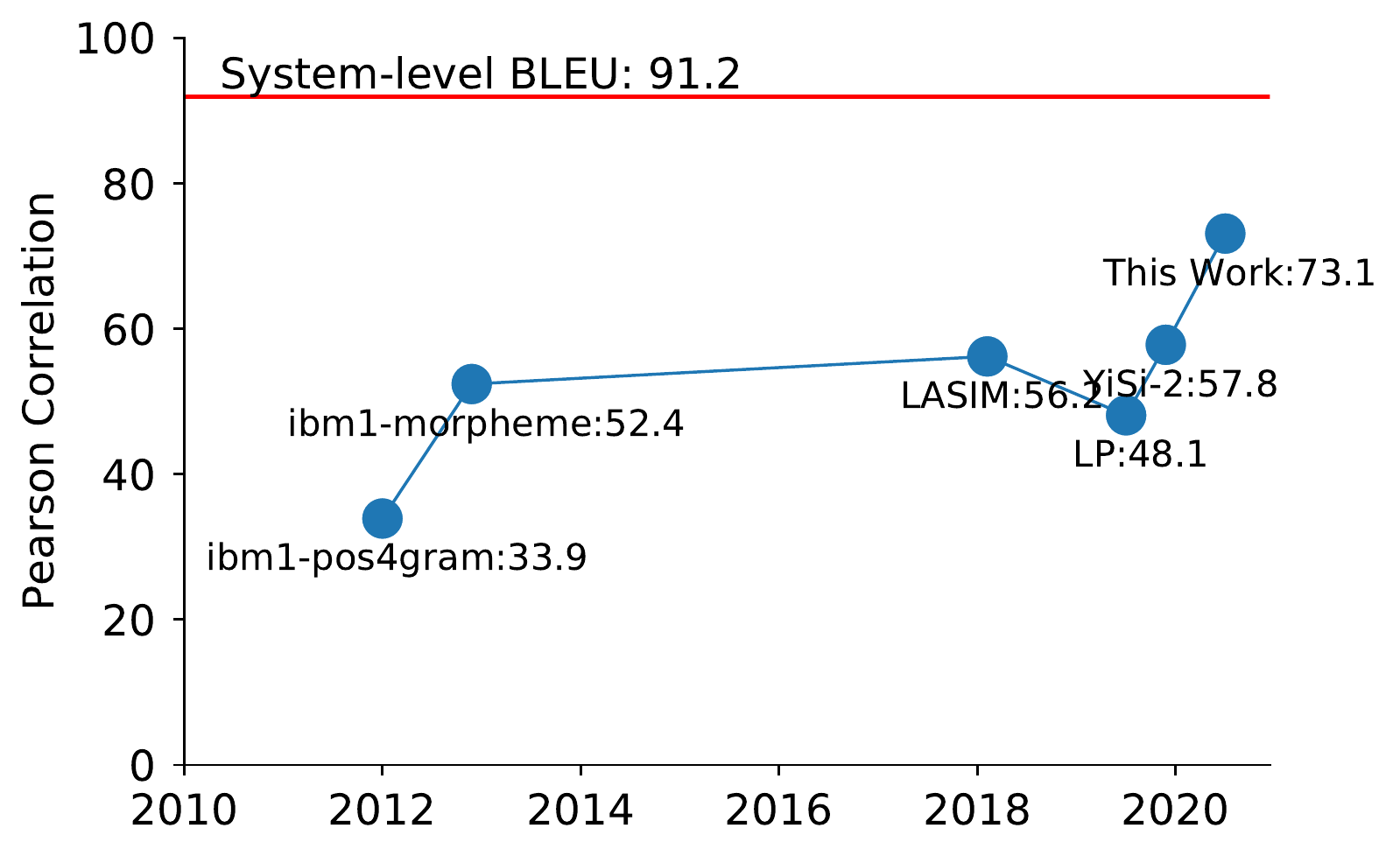}
\caption{Average results of our metric best-performing metric, together with the official results of reference-free metrics, and reference-based BLEU on system-level WMT19. 
% All are reference-free metrics except the BLEU baseline. Results are averages over six language pairs.
}
\label{fig:wmt_19_sys}
\end{figure}
Figure \ref{fig:wmt_17_sys_seg} shows that our metric %, namely 
\metric{Mover-2 + CLP(M-BERT) $\oplus$ LM}, operating on modified M-BERT with the post-hoc re-mapping and combining a target-side LM, 
% \todo{SE: which one is this?}
% Our best-performing metrics 
outperforms BLEU \Wei{by 5.7 points} in segment-level evaluation and achieves comparable performance %when contrasting them against
in the system-level evaluation. %judgments.
% We almost reach the 
% % \emph{supervised} 
% reference-based metric CHRF++. 
% In Table \ref{tab:wmt19-to-en-sys}, 
Figure \ref{fig:wmt_19_sys} shows that the same metric obtains 15.3 points gains (73.1 vs.\ 57.8), averaged over 7 languages, %results 
on WMT19 (system-level) compared to the the state-of-the-art reference-free metric YiSi-2. %Indeed, the results on all language pairs except gu-en perform rather good, comparable with system-level BLEU; see Table \ref{tab:wmt19-to-en-sys} in the appendix.
Except for one language pair, gu-en, our metric performs on a par with the reference-based BLEU (see Table \ref{tab:wmt19-to-en-sys} in the Appendix) on system-level.

% that the same metric improves over the %state-of-the-art, 
% likewise reference-free state-of-the-art YiSi-2 by on average \Wei{28 points (86.4 vs.\ 58.1)} on WMT19 \se{at system-level}.
%\Wei{we compare the proposed metrics to six unsupervised reference-free metrics.}
%including the state-of-the-art YiSi and two variants of DCU \cite{fonseca-etal-2019-findings}, %so called 
%ibm1-morpheme and ibm1-pos4gram.}
% we show that our metric operating on M-BERT and CLP alignment %can push the correlation of reference-free metrics from 57.8 to 73.1 on average on WMT19 system-level dataset. 
% improves over the likewise reference-free YiSi-2 by \Wei{28 points (86.4 vs.\ 58.1)} on average over six lanugage pairs in WMT19.
% with our correlation with humans being about 16 points higher (74.2 vs.\ 57.8). 

% there are less than 100MB sizes of pretraining
% no large pretraining corpus, e.g., less than 100MB size of 
\Wei{%For a rigorous metrics comparison, 
In Table \ref{tab:wmt17-to-en-seg}, 
%we %exhaustively 
%compare %our metrics results for a number of reference-free metrics proposed in recent years.}
we exhaustively 
compare results for several of our metric variants, based either on M-BERT or LASER. 
}
% \ref{tab:wmt17-to-en-sys} compare reference-free metrics with
% segment- and system-level human judgments . %In all language pairs, 
\se{%We can clearly see that 
%We observe that the 
%Our metric which operates on M-BERT modified by post-hoc alignment achieves 11 points (34.0 vs.\ 45.0) better results overall than counterparts based on original M-BERT, while the same modification to LASER shows less improvements. 
%\Wei{It appears that the metric based on misaligned M-BERT, Mover-1 + M-BERT, still outperforms the counterpart with re-mapped static embedding RCSLS, by a large margin.}
% \Wei{We note that our metric based on original M-BERT outperforms the aligned static embedding RCSLS by 5.2 points}
We note that re-mapping has considerable effect for M-BERT (up to 10 points  improvements), but much less so for LASER.} \se{We believe that this is because the underlying embedding space of LASER is less `misaligned' since it has been (pre-)trained on parallel data.\footnote{\Wei{However, in the appendix, we find that re-mapping LASER using 2k parallel sentences achieves considerable improvements on low-resource languages, e.g., kk-en (from -61.1 to 49.8) and lt-en (from 68.3 to 75.9); see Table \ref{tab:wmt19-to-en-sys}.}
% re-mapping LASER may be beneficial for language pairs with very little parallel data,  
% e.g., kk-en (from -0.61 to 0.50) and lt-en (from 0.68 to 0.75); see Table \ref{tab:wmt19-to-en-sys}.
}} 
% \todo{MP: Broken ref here.} 
%(Appendix).} %\todo{SE: why are these not reported in the main text?} 
% our best metrics, operating on cross-lingual word embeddings (M-BERT) modified by post-hoc alignments %can substantially 
% outperform the metrics based on aligned sentence embeddings (LASER)--however, before re-mapping, LASER is overall better, indicating that M-BERT suffers from cross-lingual misalignments much more strongly\Wei{--as LASER is pretrained on parallel corpora over languages.}
% \todo{SE: LASER is better. Only after re-mapping is MBERT better.}
%However, the gap closes after combining with the language model.
% \todo{Even after re-mapping two MBERTs are worse than two LASERs.}
%We believe that the improvement comes from post-hoc alignments that can address the mismatch in semantics between cross-lingual embeddings, and adding language modeling on the target side can help to penalize literal translations (see the discussion in \S\ref{sec:W2W}). 
\se{
While the re-mapping %thus has a considerable effect 
is thus effective 
for metrics based on M-BERT, %it is still required to combine perplexity scores yielded by target-side LM for surpassing BLEU.}
we still require the target-side LM to outperform BLEU. 
}
% outperform BLEU only after combining the metrics with a target-side LM. 
We assume the LM can address challenges that the re-mapping apparently is not able to %properly 
handle properly; see our discussion in \S\ref{sec:W2W}. 
%\Wei{\subsection{What Combination is Good} %We obverse that 

Overall, we remark that none of our metric combinations %win over language pairs and WMT datasets. 
performs consistently best. \se{The reason may be that LASER and M-BERT are pretrained over hundreds of languages with substantial differences in %the corpus sizes, which we assume play a role, 
corpora sizes 
in addition to the different effects of the re-mapping.} 
% We speculate that %varying 
% different sizes of pretraining data over languages %while
% for LASER and M-BERT may play a role, in addition to the different effects of re-mapping. 
%for LASER and M-BERT. 
However, we observe that \metric{Mover-2 + CLP(M-BERT)} %consistently 
performs best on average over all language pairs when the LM is not added. %Yet, with 
When 
the LM is added, \metric{Mover-2 + CLP(M-BERT) $\oplus$ LM} and \metric{Cosine + UMD (LASER) $\oplus$ LM} perform comparably. %We speculate 
This indicates that there may be a saturation effect when \Wei{it comes to the LM} or that the LM coefficients should be tuned individually %over M-BERT and LASER.
for each semantic similarity metric based on cross-lingual representations.

%% file: tables/wmt17-segment.tex
% \setlength{\tabcolsep}{4.5pt}
\begin{table*}[h!]
    \setlength{\tabcolsep}{3.3pt}
    \footnotesize    
    \centering
    \begin{tabular}{l | l ccccccc c}
    \toprule
    % & & \multicolumn{8}{c}{\textbf{Direct Assessment}}\\
    Setting & Metrics & cs-en & de-en & fi-en & lv-en & ru-en & tr-en & zh-en & Average \\
    
    \midrule
    \multirow{2}{*}{$m(\mathbf{y^*},\mathbf{y})$}
    &\metric{sentBLEU} & 43.5 & 43.2 & 57.1 & 39.3 & 48.4 & 53.8 & 51.2 & 48.1 \\
    % &\metric{METEOR++} & 0.552 & 0.538 & 0.720 & 0.563 & 0.627 & 0.626 & 0.646 & 0.610 \\
    &\metric{chrF++} & \textbf{52.3} & \textbf{53.4} & \textbf{67.8} & \textbf{52.0} & \textbf{58.8} & \textbf{61.4} & \textbf{59.3} & \textbf{57.9} \\
    % &\metric{Mover-2 + BERT} & \textbf{0.679} & \textbf{0.710} & \textbf{0.832} & \textbf{0.745} & \textbf{0.736} & \textbf{0.763} & \textbf{0.740} & \textbf{0.743} \\
    \midrule
    \multirow{17}{*}{$m(\mathbf{x},\mathbf{y})$} 
    &\multicolumn{9}{l}{\textit{Baseline with Original Embeddings}}\\
    \cmidrule{2-10}
    &\metric{Mover-1 + M-BERT} & 22.7 & 37.1 & 34.8 & 26.0 & 26.7 & \textbf{42.5} & \textbf{48.2} & 34.0 \\
    &\metric{Cosine + LASER} & \textbf{32.6} & \textbf{40.2} & \textbf{41.4} & \textbf{48.3} & \textbf{36.3} & 42.3 & 46.7 & \textbf{41.1} \\
    \cmidrule{2-10}
    &\multicolumn{9}{l}{\textit{Cross-lingual Alignment for Sentence Embedding}}\\
    \cmidrule{2-10}
    &\metric{Cosine + CLP(LASER)} & 33.4 & \textbf{40.5} & 42.0 & \textbf{48.6} & \textbf{36.0} & 44.7 & 42.2 & \textbf{41.1} \\
    &\metric{Cosine + UMD(LASER)} & \textbf{36.6} & 28.1 & \textbf{45.5} & 48.5 & 31.3 & \textbf{46.2} & \textbf{49.4} & 40.8 \\
    % &\metric{Cosine + UMD $\circ$ CLP(LASER)} & 0.322 & 0.226 & 0.428 & 0.458 & 0.295 & 0.453 & 0.449 & 0.376 \\
    % &\metric{Cosine + CLP $\circ$ UMD(LASER)} & 0.345 & 0.415 & 0.452 & \textbf{0.486} & \textbf{0.422} & 0.457 & 0.430 & \textbf{0.430} \\
    
    \cmidrule{2-10}
    &\multicolumn{9}{l}{\textit{Cross-lingual Alignment for Word Embedding}}\\
    \cmidrule{2-10}
    &\metric{Mover-1 + RCSLS} & 18.9 & 26.4 & 31.9 & 33.1 & 25.7 & 31.1 & 34.3 & 28.8\\
    &\metric{Mover-1 + CLP(M-BERT)} & 33.4 & 38.6 & 50.8 & 48.0 & 33.9 & \textbf{51.6} & 53.2 & 44.2 \\
    &\metric{Mover-2 + CLP(M-BERT)} & \textbf{33.7} & 38.8 & \textbf{52.2} & \textbf{50.3} & \textbf{35.4} & 51.0 & \textbf{53.3} & \textbf{45.0} \\
    &\metric{Mover-1 + UMD(M-BERT)} & 22.3 & 38.1 & 34.5 & 30.5 & 31.2 & 43.5 & 48.6 & 35.5 \\
    &\metric{Mover-2 + UMD(M-BERT)} & 23.1 & 38.9 & 37.1 & 34.7 & 33.0 & 44.8 & 48.9 & 37.2 \\
    % &\metric{Mover-1 + UMD $\circ$ CLP(M-BERT)} & 0.269 & 0.395 & 0.449 & 0.357 & 0.323 & 0.515 & 0.535 & 0.406 \\
    % &\metric{Mover-1 + CLP $\circ$ UMD(M-BERT)} & 0.331 & 0.391 & 0.506 & 0.477 & 0.339 & \textbf{0.517} & 0.530 & 0.442 \\
    % &\metric{Mover-2 + CLP $\circ$ UMD(M-BERT)} & 0.335 & \textbf{0.395} & 0.521 & 0.501 & \textbf{0.355} & \textbf{0.512} & 0.531 & \textbf{0.450} \\
    \cmidrule{2-10}
    &\multicolumn{9}{l}{\textit{Combining Language Model}}\\
    \cmidrule{2-10}
    &\metric{Cosine + CLP(LASER) $\oplus$ LM} & 48.8 & 46.7 & 63.2 & 66.2 & 51.0 & 54.6 & 48.6 & 54.2\\
    &\metric{Cosine + UMD(LASER) $\oplus$ LM} & \textbf{49.4} & 46.2 & \textbf{64.7} & 66.4 & \textbf{51.1} & \textbf{56.0} & 52.8 & \textbf{55.2}\\
    &\metric{Mover-2 + CLP(M-BERT) $\oplus$  LM} & 46.5 & 46.4 & 63.3 & 63.8 & 47.6 & 55.5 & \textbf{53.5} & 53.8 \\
    &\metric{Mover-2 + UMD(M-BERT) $\oplus$  LM} & 41.8 & \textbf{46.8} & 60.4 & 59.8 & 46.1 & 53.8 & 52.4 & 51.6 \\
    % &\metric{Cosine + UMD $\circ$ CLP(LASER) + LM} & 0.492 & 0.444 & \textbf{0.649} & 0.654 & 0.504 & 0.536 & \textbf{0.539} & 0.545\\
    % &\metric{Cosine + CLP $\circ$ UMD(LASER) + LM} & \textbf{0.491} & \textbf{0.473} & 0.640 & \textbf{0.662} & \textbf{0.533} & 0.550 & 0.491 & \textbf{0.549}\\
    % &\metric{Mover-1 + CLP $\circ$ UMD(M-BERT) + LM} & 0.467 & 0.459 & 0.633 & 0.630 & 0.470 & \textbf{0.564} & 0.535 & 0.537 \\
    % &\metric{Mover-2 + CLP $\circ$ UMD(M-BERT) + LM} & 0.461 & 0.458 & 0.632 & 0.632 & 0.474 & 0.549 & 0.540 & 0.535 \\
	\bottomrule  
    \end{tabular}
    \caption{Pearson correlations with segment-level human judgments on the WMT17 dataset. 
    \label{tab:wmt17-to-en-seg} }
% \vspace{-0.1in}
\end{table*}

%% file: 4-analysis.tex
\section{Analysis}
\input{tables/analysis_examples}

We first analyze preferences of our metrics based on M-BERT and LASER (\S\ref{sec:W2W}) and then examine how much parallel data we need for re-mapping our vector spaces (\S\ref{sec:size}). Finally, we discuss whether it is legitimate to correlate our metric scores, which evaluate the similarity of system predictions and source texts, to human judgments based on system predictions and references (\S\ref{sec:human}). 

\subsection{Metric preferences}\label{sec:W2W}

\insertDistplots

\se{To analyze %our metrics based on M-BERT and LASER and in particular 
why our metrics based on M-BERT and LASER  perform so badly for the task of reference-free MT evaluation, we query them for their preferences. In particular, for a fixed source sentence $\mathbf{x}$,   we consider two target sentences  $\tilde{\mathbf{y}}$ and $\hat{\mathbf{y}}$ and %we consider 
evaluate the following score difference:
\begin{align}\label{eq:pref}
    d(\tilde{\mathbf{y}},\hat{\mathbf{y}}; \mathbf{x}):=m(\mathbf{x},\tilde{\mathbf{y}}) - m(\mathbf{x},\hat{\mathbf{y}})
\end{align}
}
When $d>0$, then metric %\footnote{For $m$, we choose the cosine distance for LASER and the cross-lingual variant of MoverScore for M-BERT.} 
$m$ prefers $\tilde{\mathbf{y}}$ over $\hat{\mathbf{y}}$, given $\mathbf{x}$, and when $d<0$, this relationship is reversed. 
In the following, we %consider preferences of our metrics of a s 
compare preferences of our metrics for specifically modified target sentences $\tilde{\mathbf{y}}$ over the human references $\mathbf{y}^{\star}$. We choose $\tilde{\mathbf{y}}$ to be (i) a random reordering of $\mathbf{y}^{\star}$, to ensure that our metrics do not have the BOW (bag-of-words) property, (ii) a word-order preserving translation of $\mathbf{x}$, %i.e., %human-written or automatic word-by-word translation 
%coupled with human reordering 
%of the English $\mathbf{y}^{\star})$.
i.e., (ii-a) %expert and automatic 
an expert reordering of the human $\mathbf{y}^{\star}$ to have the same word order as $\mathbf{x}$ 
as well as 
(ii-b) 
a word-by-word translation, obtained either using experts or automatically. 
Especially condition (ii-b) %especially 
tests for preferences for literal translations, a common MT-system property. 

%\Wei{Without the LM combined, our reference-free metrics, operating on M-BERT, cannot outperform the counterpart baseline, reference-based BLEU.}
%\Wei{The re-mapping solution 
%apparently mitigate word-level misalignment  in the underlying embedding space, however, it appears to have the inability of handling a recurring MT challenge, namely translationese.}

\se{%To investigate preference of the original embedding spaces and our various modifications for either human reference translations or translationese, we conducted the following analysis.  
%We %used Google Translate (GT) to translate XX German sentences into English word-by-word, i.e., we %made a lexicon lookup for German words using Google. 
%In addition to that, 
%For condition (1), we randomly shuffle the English sentence $\mathbf{y}^{\star}$. %, to check whether our models are bag-of-words models. 
\paragraph{Expert word-by-word translations.}
%For 50 selected German sentences, 
We %further 
had an expert (one of the co-authors) translate 50 German sentences 
%them 
word-by-word into English.
% and we had the expert in addition reorder the English $\mathbf{y}^{\star}$ in the word order of the German source sentence $\mathbf{x}$. 
Table \ref{table:reorder} illustrates this scenario. We note how bad the word-by-word translations sometimes are even for %the 
closely related language pairs such as 
German-English. For example, the word-by-word translations in English retain the original German verb final positions, leading to quite ungrammatical English translations.  %resulting in grammatically very bad English constructions. 
}

%For example, the sentence \emph{``Putin divided out and accused Ankara Russia in the move like to his''} is given by the word-by-word translation, while the reference translation $y^\star$ is \emph{``Mr Putin lashed out accusing Ankara of stabbing Moscow in the back''}. 

\se{Figure \ref{fig:coverage} shows histograms for the $d$ statistic for the 50 selected sentences. 
We first check condition (i) for the 50 sentences. 
%This illustrates that %(1) 
We observe that 
both %M-BERT, combined with MoverScore, and LASER 
\metric{Mover + M-BERT} and \metric{Cosine+LASER}
prefer the original human references over random reorderings, indicating that they %indeed encode syntax. 
are not BOW models, a reassuring finding. 
Concerning (ii-a), they are largely indifferent between correct English word order and the situation where the word order of the human reference is %reordered to German syntax.
the same as the German. %(2b). 
Finally, they strongly prefer the expert word-by-word translations over the human references (ii-b). 
%but they are indifferent between human reference and GT word-by-word translation. %For GT, they are again indifferent, indicating that the GT word-by-word translations are of inferior quality.
}

%\se{Taken together, %our analysis indicates the following: 
%this yields the following conclusions: 
%(i) we reassuringly that %our metrics based on 
%BERT and LASER do not have the BOW property; 
%(i) it seems that %they 
%M-BERT and LASER 
%are mostly indifferent between correct target language word order and the situation where source and target language have identical word order; (ii) this appears 
%(2b) in part explains why our metrics
%to make them 
%this 
\se{Condition (ii-a) in part 
explains why our metrics 
prefer expert word-by-word translations the most: for a given source text, these have higher lexical overlap than human references and, by (ii-a), %in addition 
they have a favorable target language syntax, \emph{viz.}, where the source and target language word order are equal. 
%This %explains 
Preference for translationese, (ii-b), in turn 
is apparently a main reason 
why our metrics do not perform well, by themselves and without a language model, as reference-free MT evaluation metrics. More worryingly, it indicates that cross-lingual M-BERT and LASER are not robust to the `adversarial inputs' given by MT systems. 
%It also indicates that %, while 
%the current practice of evaluating them on zero-shot text classification tasks is not adequate for %testing 
%assessing 
%more fine-grained and nuanced cross-lingual text  understanding.}
}

\iffalse
To investigate the impact of this phenomena in WMT evaluation, we report its coverage in three language pairs (cs-en, de-en and fi-en) on WMT17 benchmark. 
% Due to the page constraints, we take three language pairs (cs-en, de-en and fi-en) as prime examples. 
Figure \ref{fig:coverage} illustrates that all language pairs contains a large amount of word-by-word translations.
\fi
% We suspect that cross-lingual alignments trained on precisely matched word and sentence pairs can address the semantic mismatch for multilingual embeddings in the shared embedding space, however, syntactic information like word order is not required to learn such alignments, resulting in the elusiveness of whether word-by-word translations can be well-identified. 

\paragraph{Automatic word-by-word translations.} 
%In addition, we observe that the GT word-by-word translations are of even lower quality since they often pick the wrong word senses---e.g., \emph{his} as a translation of German \emph{sein}, which depending on context may %mean \emph{like} or \emph{has fallen}.
%be a personal pronoun or the infinitive \emph{to be}. 

For a large-scale analysis of condition (ii-b) across different language pairs, we resort to automatic word-by-word translations obtained from Google Translate (GT). 
%We 
%compile a German-English word-level dictionary from Google Translate (GT) to translate XX German sentences into English word-by-word. 
To do so, we %look up each word 
go over each word in the source sentence $\mathbf{x}$ from left to right, look up its translation in GT \emph{independently of context} %in GT 
and replace the word by the obtained translation. %to compile dictionaries. 
When a word has several translations, we keep the first one offered by GT. 
Due to context-independence, 
%we observe that 
the GT word-by-word translations are of much lower quality than the expert word-by-word translations since they often pick the wrong word senses---e.g., %\emph{his} as a translation of German \emph{sein}, which depending on context may %mean \emph{like} or \emph{has fallen}.
the German word \emph{sein} may either be a personal pronoun (\emph{his}) or the infinitive \emph{to be}, which would be selected correctly only by chance; cf.\ Table \ref{table:reorder}. 

Instead of reporting histograms of $d$, we define a ``W2W'' statistic that counts the relative number of times that $d(\mathbf{x}',\mathbf{y}^\star)$ is positive, where $\mathbf{x}'$ denotes the described literal translation of $\mathbf{x}$ into the target language:
\begin{align}\label{eq:w2w}
    \text{W2W}:=\frac{1}{N}\sum_{(\mathbf{x}',\mathbf{y}^\star)} I(\,d(\mathbf{x}',\mathbf{y}^\star)>0\,)
\end{align}
Here $N$ normalizes W2W to lie in $[0,1]$ and a high W2W score indicates the metric prefers translationese over human-written references.  
Table \ref{tab:translationese} shows that reference-free metrics with %or
original embeddings (LASER and M-BERT) either still prefer literal over human translations (e.g., W2W score of 70.2\% for cs-en) or struggle in distinguishing them. %---as 
%, to which human often dislike and assign low scores.
% That metrics, operating on LASER and 
%(but also, to a lesser degree, re-mapped) 
% M-BERT, prefer such bad translations over human reference translations 
%This exposes an important and severe deficiency of the embedding space. 
%\Wei{%With re-mapped embeddings, our metrics are still indifferent between literal translations and gold references, indicating that re-mapping handles translationese to a marginal degree.}
% Our re-mapping to some helps considerably for M-BERT, but overall our metrics even the re-mapped metric is almost indifferent between literal translations and gold references. We think that this shows that re-mapping may address some syntactic phenomena as well, 
%(as BERT stores syntax trees \citep{hewitt-manning-2019-structural}, the re-mapped representations may )
% but %the main benefit of re-mapping is semantic in nature.
% to a marginal degree. 
% To this end, we probe how reference-free metrics can be fooled by word-by-word translation. 
% we find that reference-free metrics can be fooled by word-by-word translation. To this end, we synthetically create word-by-word translations of the source sentence using Google Translate
% , and directly use the source sentence as translation to create full code-switching translation. 
\se{Re-mapping helps to a small degree.} 
\se{Only} when combined with the LM scores do we get adequate scores for the W2W statistic. 
Indeed, the LM is expected to capture unnatural word order in the target language and penalize word-by-word translations by recognizing them as much less likely to appear in a language.

Note that for expert word-by-word translations, we would expect the metrics to  perform even worse. 

\iffalse
As expected, Table \ref{tab:translationese} shows that our metrics with cross-lingual alignment  
achieve small improvement but are still indifferent from literal and human translation (about 0.4 in Mover-based metrics). This indicates that BERT can encode syntax tree as discussed in \cite{hewitt-manning-2019-structural}, which can help to penalize literal translation, however, experiments show this effect is indeed weak. 
\fi 

\begin{table}[h!]
    \setlength{\tabcolsep}{3.2pt}
    \footnotesize    
    \centering
    \begin{tabular}{@{}l | r rr@{}}
    \toprule
    % & & \multicolumn{8}{c}{\textbf{Direct Assessment}}\\
    Metrics & cs-en & de-en & fi-en  \\
    \midrule
    % \cmidrule{2-10}
    % &\multicolumn{2}{l}{\textit{Before Cross-lingual Alignment}}\\
    % \cmidrule{2-10}
    % \multirow{5}{*}{$m(\mathbf{x},\mathbf{y})$} 
    \metric{Cosine + LASER} & 70.2 & 65.7 & 53.9 \\
    \metric{Cosine + CLP(LASER)} & 70.7 & 64.8 & 53.7 \\
    \metric{Cosine + UMD(LASER)} & 67.5 & 59.5 & 52.9 \\
    \metric{Cosine + UMD(LASER) $\oplus$ LM} & \textbf{7.0} & \textbf{7.1} & \textbf{6.4} \\
    \midrule
    \metric{Mover-2 + M-BERT} & 61.8 & 50.2 & 45.9  \\
    \metric{Mover-2 + CLP(M-BERT)} & 44.6 & 44.5 & 32.0 \\
    \metric{Mover-2 + UMD(M-BERT)} & 54.5 & 44.3 & 39.6\\
    \metric{Mover-2 + CLP(M-BERT) $\oplus$ LM} & \textbf{7.3} & \textbf{10.2} & \textbf{6.4} \\
	\bottomrule  
    \end{tabular}
    \caption{W2W statistics for selected language pairs. Numbers are in percent.}
    \label{tab:translationese}
\vspace{-0.1in}
\end{table}
\subsection{\se{Size of} Parallel Corpora}\label{sec:size}
% \begin{figure}
%  \begin{subfigure}[b]{0.5\linewidth}
%     \centering
%     \includegraphics[width=\linewidth]{figures/cs_en_training_size.pdf}
%     \label{fig7:a} 
%     \vspace{-0.15in}
%   \end{subfigure}%% 
%   \begin{subfigure}[b]{0.5\linewidth}
%     \centering
%     \includegraphics[width=\linewidth]{figures/de_en_training_size.pdf} 
%     \label{fig7:b} 
%     \vspace{-0.15in}
%   \end{subfigure} 
%   \begin{subfigure}[b]{0.5\linewidth}
%     \centering
%     \includegraphics[width=\linewidth]{figures/fi_en_training_size.pdf}
%     \vspace{-0.2in}
%   \end{subfigure}%%
%   \begin{subfigure}[b]{0.5\linewidth}
%     \centering
%     \includegraphics[width=\linewidth]{figures/lv_en_training_size.pdf}
%     \vspace{-0.2in}
%   \end{subfigure} 
%   \caption{Comparison in sentence and word alignment across debiasing methods, where x-axis denotes the size of parallel corpus for training alignment across languages.}
%   \label{fig:training_size}
% \vspace{-0.15in}
% \end{figure} 
\begin{figure}[t]
\centering
\includegraphics[width=\linewidth]{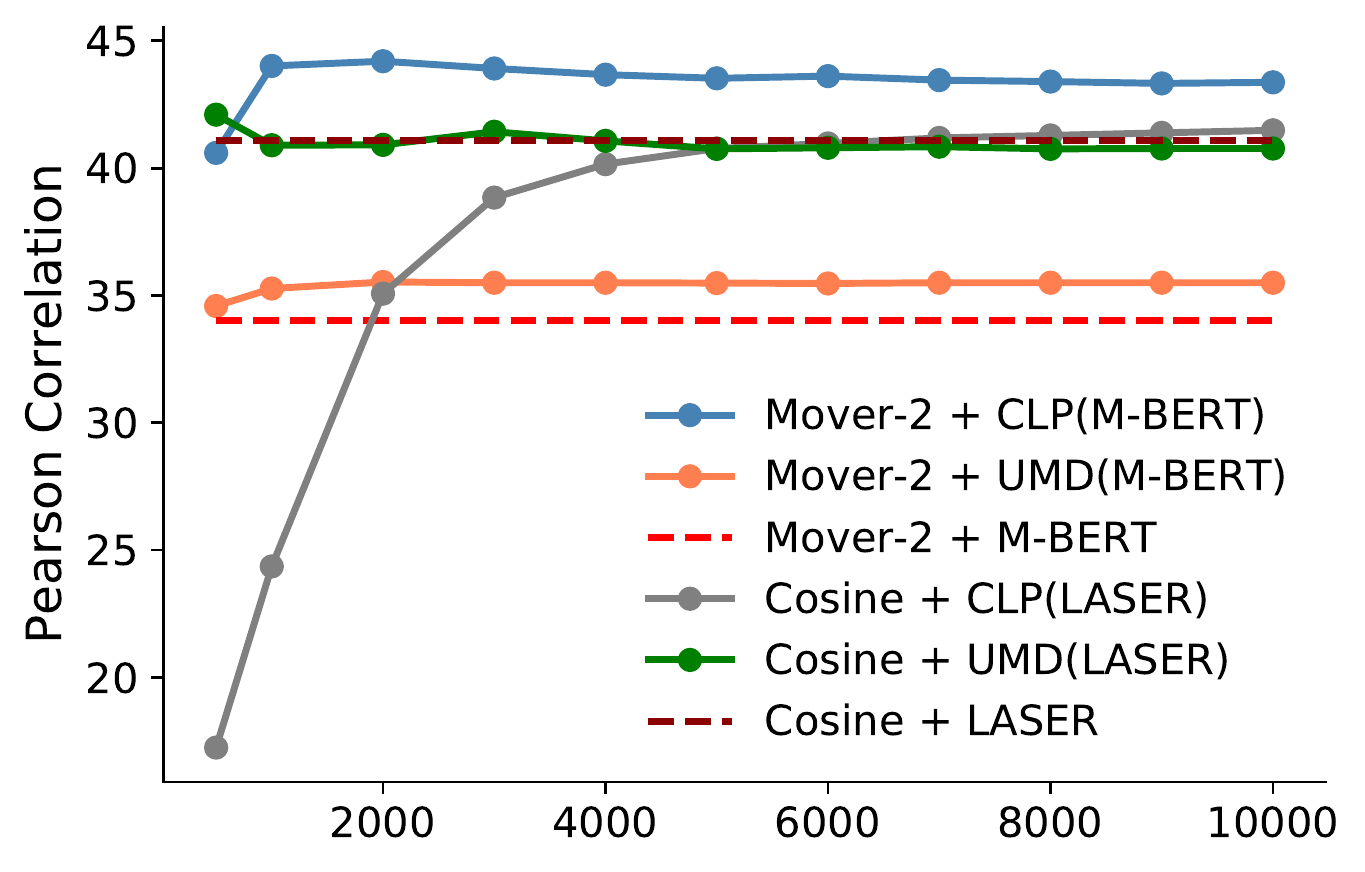}
\caption{Average results of our metrics based on sentence- and word-based re-mappings of vector spaces %alignments %which are trained on 
as a function of different sizes of parallel corpus (x-axis).}
 \label{fig:training_size}
\vspace{-0.15in}
\end{figure} 
Figure \ref{fig:training_size} compares sentence- and word-level re-mapping trained with a varying number of parallel sentences. 
%Overall, metrics building on sentence embeddings (e.g., Cosine + LASER) outperform metrics (e.g., Mover-2 + M-BERT) based on word embedding.\todo{SE: What the hack?!?! Before you claimed the opposite} However, the latter (e.g., Mover-2 + CLP(M-BERT)) 
Metrics based on M-BERT result in the highest correlations after re-mapping, even with a small amount of training data (1k).
% LASERScore\todo{SE: LASERScore is not defined..} outperforms the word-level metric MoverScore.  However, CLP(MoverScore) results in the highest correlation after cross-lingual word alignment, even with small amount of training data (1k) while sentence-level mapping often underperforms. 
We observe that \metric{Cosine + CLP(LASER)} and \metric{Mover-2 + CLP(M-BERT)} show very similar trends with a sharp increase with increasing amounts of parallel data and then level off quickly. 
%in the early stage and saturation afterwards. 
However, the M-BERT based Mover-2 reaches its peak and outperforms the original baseline with only 1k data, while LASER needs 2k before beating the corresponding original baseline. 
%Indeed, Mover-2 + CLP(M-BERT) is at the word-level and many more word pairs can be extracted from sentence pairs.
%However, the former can outperform its corresponding metric without re-mapping using about 1k training data while the latter needs 2k, since one can extract massive word pairs from 1k sentence pairs.

% from which we find that they can outperform corresponding metrics without re-mapping embeddings, using 1-2k training data. Compared to 

% first 1k-4k parallel corpus\todo{SE: `in the first ... parallel corpus' - what does that mean?} while UMD stays mildly against the size of corpus with less  performance gains than CLP.\todo{SE: stays mildly against ... is not grammatically correct}

% Results are averaged on 5 language pairs, excluded Chinese, because it is hard to make literal translation with Google Translation API due to the mismatch between word segmentations.
% cs-en, de-en, fi-en, ru-en and tr-en excluded Chinese 

% The lower score means that CL metrics are less fooled by word-to-word translation.

% In a short summary, MoverScore is less preferred to fully word-to-word translation than LASERScore, and CLP(MoverScore) can further debias this preference, however, MoverScore shows stronger preference in fully code-switching translation than LASERScore. These observations indicate that there is no clear winner in both cases…

\subsection{Human Judgments}\label{sec:human}
%For each system translation, 
%we collect 
The WMT datasets contain
segment- and system-level human judgments that we use for evaluating the quality of our reference-free metrics. The segment-level judgments assign one direct assessment (DA) score to each pair of system and human translation, while system-level judgments associate each system with a single DA score averaged across all pairs in the dataset.
% assigning one direct assessment (DA) score to the system translation against reference.
%However, a cross-lingual DA score, which we will call CLDA, which compares system translations against source is not provided. 
We initially suspected the DA scores to be biased for our setup---which compares $\mathbf{x}$ with $\mathbf{y}$---as they are based on comparing $\mathbf{y}^\star$ and $\mathbf{y}$. 
Indeed, it is known that (especially) human professional translators ``improve'' $\mathbf{y}^\star$, e.g., by making it more readable, relative to the original $\mathbf{x}$ \citep{rabinovich-etal-2017-found}.  %\todo{MP: it seems we have a missing citation here} 
%Indeed, we do find such cases in our data. To address this issue, we collected human cross-lingual DA (CLDA) assessments of $(\mathbf{x},\mathbf{y})$ for WMT17 and three language pairs (de-en, ru-en, zh-en), following %similar  
%guidelines akin to those for the DA scores. However, we found that our annotators had similar agreement levels with each other as they had with the DA scores. As we collected comparably few CLDA scores, we therefore decided to use the DA scores, treating them as reliable scores for our setup.   
We investigated the validity of DA scores by collecting human assessments in the cross-lingual settings (CLDA), where annotators directly compare source and translation pairs $(\mathbf{x},\mathbf{y})$ from the WMT17 dataset.
This small-scale manual analysis hints that DA scores are a valid proxy for CLDA. Therefore, we decided to treat them as reliable scores for our setup and evaluate our proposed metrics by comparing their correlation with DA scores.

%% file: tables/analysis_examples.tex
\begin{table*}[!htb]
    \centering
    {\small
    \begin{tabular}{l|lll} \toprule
         $\mathbf{x}$ &  Dieser von Langsamkeit gepr{\"a}gte Lebensstil scheint aber ein Patentrezept f{\"u}r ein hohes Alter zu sein.\\
         $\mathbf{y}^{\star}$& However, this slow pace of life seems to be the key to a long life.\\
         $\mathbf{y}^{\star}$-random & To pace slow seems be the this life. life to a key however, of long\\
         $\mathbf{y}^{\star}$-reordered & This slow pace of life seems however the key to a long life to be.\\
         $\mathbf{x}'$-GT & This from slowness embossed lifestyle seems but on nostrum for on high older to his.\\ 
         $\mathbf{x}'$-expert & This of slow pace characterized life style seems however a patent recipe for a high age to be.\\ \midrule
         $\mathbf{x}$ & Putin teilte aus und beschuldigte Ankara, Russland in den R{\"u}cken gefallen zu sein.\\
         $\mathbf{y}^\star$ & Mr Putin lashed out, accusing Ankara of stabbing Moscow in the back.\\
         $\mathbf{y}^{\star}$-random & Moscow accusing lashed Putin the in Ankara out, Mr of back. stabbing\\
         $\mathbf{y}^{\star}$-reordered & Mr Putin lashed out, accusing Ankara of Moscow in the back stabbing.\\
         $\mathbf{x}'$-GT & Putin divided out and accused Ankara Russia in the move like to his.\\
         $\mathbf{x}'$-expert & Putin lashed out and accused Ankara, Russia in the back fallen to be.\\
         \bottomrule
    \end{tabular}
    }
    \caption{Original German input sentence $\mathbf{x}$, together with the human reference $\mathbf{y}^\star$, in English, and a randomly ($\mathbf{y}^{\star}$-random) and expertly reordered ($\mathbf{y}^{\star}$-reordered) English sentence as well as expert word-by-word translation ($\mathbf{x}'$) of the German source sentence. The latter is either obtained by the human expert or by Google Translate (GT).}
    \label{table:reorder}
\end{table*}

\newcommand{\insertDistplots}{
\begin{figure*}
\begin{minipage}{0.33\textwidth}  
	\centerline{\includegraphics[width=\linewidth]{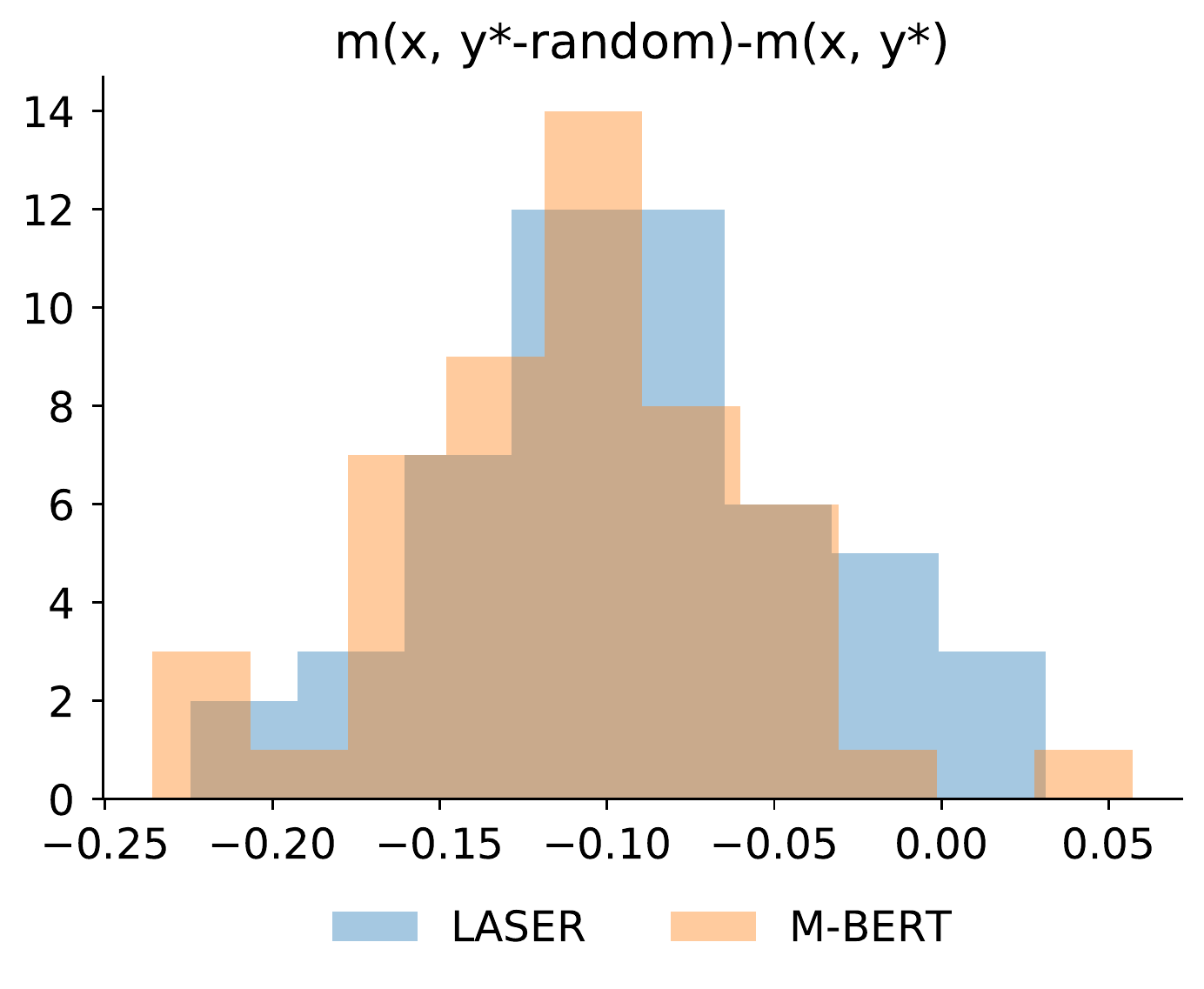}}  
\end{minipage} 
\begin{minipage}{0.33\textwidth}  
	\centerline{\includegraphics[width=\linewidth]{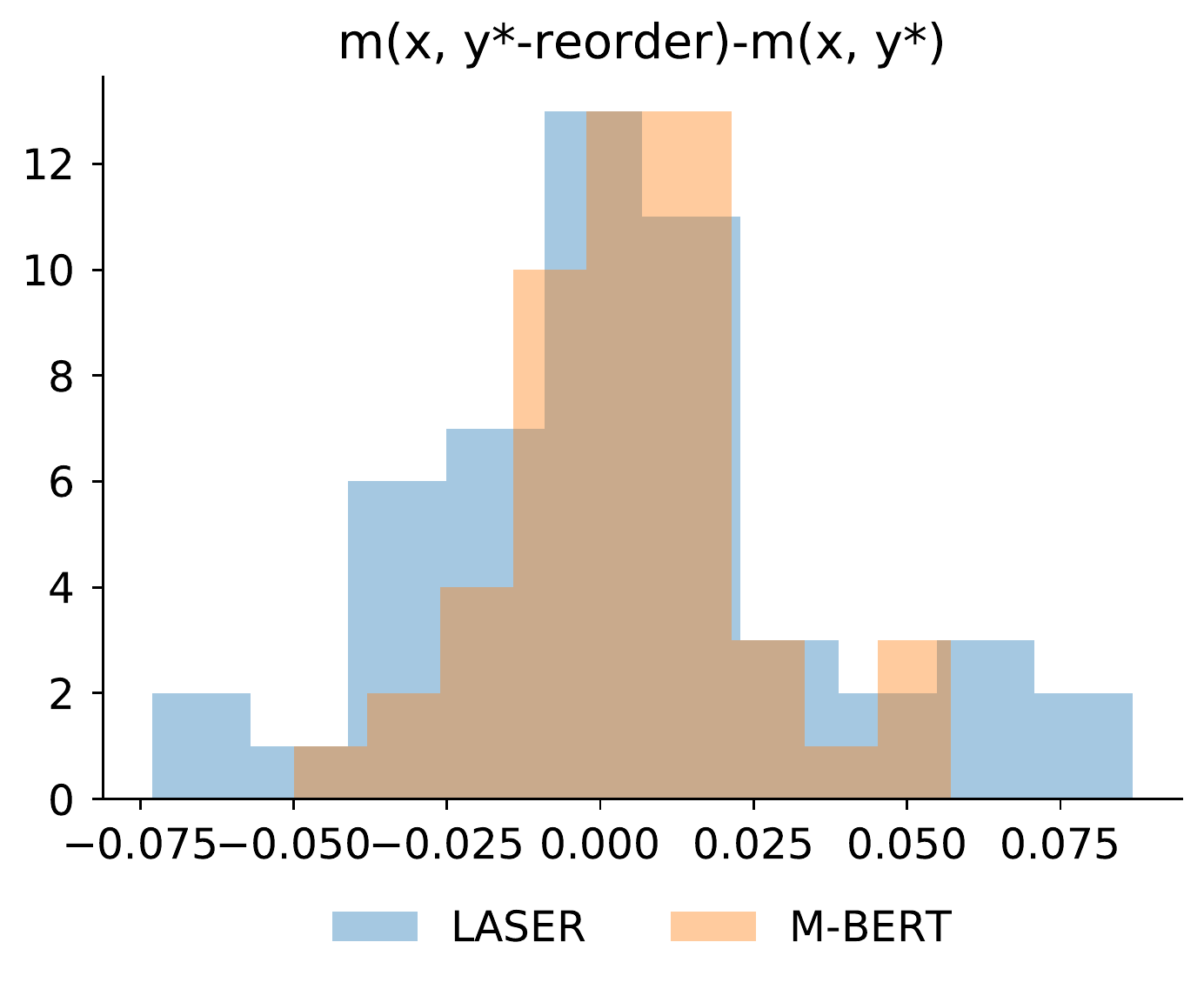}}  
\end{minipage} 
\begin{minipage}{0.33\textwidth}  
	\centerline{\includegraphics[width=\linewidth]{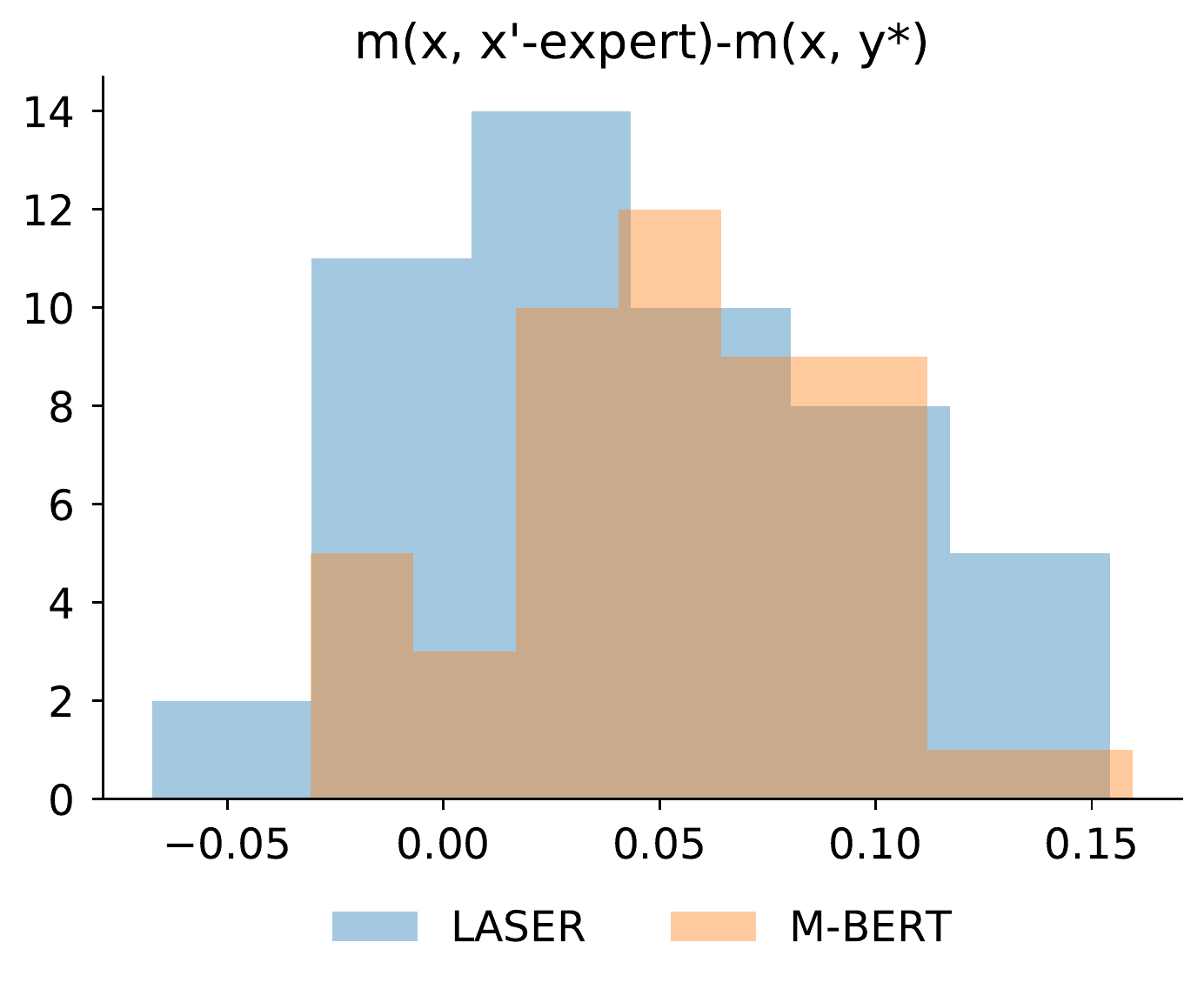}}
\end{minipage} 
\caption{Histograms of $d$ scores defined in Eq.~\eqref{eq:pref}. 
Left: Metrics based on LASER and M-BERT favor gold over randomly-shuffled human references. Middle: Metrics %hardly distinguish 
are roughly indifferent 
between gold and reordered human references. 
Right: Metrics favor expert word-by-word translations over gold human references.}
\label{fig:coverage}
\vspace{-0.15in}
\end{figure*} 
}

%% file: 5-conclusion.tex
\section{Conclusion}

Existing semantically-motivated metrics for reference-free evaluation of MT systems have so far displayed \se{rather} poor correlation with human estimates of translation quality. %\todo{SE: are we so sure about this?} 
In this work, we investigate a range of reference-free metrics based on cutting-edge models for inducing cross-lingual semantic representations: cross-lingual (contextualized) word embeddings and cross-lingual sentence embeddings. We have identified some scenarios in which these metrics fail, prominently their inability to punish literal word-by-word translations (the so-called ``translationese''). We have investigated two different mechanisms for mitigating this undesired phenomenon: (1) an additional (weakly-supervised) cross-lingual alignment step, reducing the mismatch between representations of mutual translations, and (2) language modeling (LM) on the target side, which is inherently equipped to punish ``unnatural'' sentences in the target language. 
We show that the reference-free coupling of cross-lingual similarity scores with the target-side language model surpasses the reference-based BLEU in segment-level MT evaluation.
%and considerably decreases the gap between reference-free metrics and system-level BLEU score.
% and outperforms existing reference-free metrics.

% the latter (target language LM) yields significant improvements over existing reference-free metrics and even in comparison with reference-based metrics like BLEU. 
We believe our results have two relevant implications. First, they portray the viability of reference-free MT evaluation and warrant wider research efforts in this direction. Second, they indicate that reference-free MT evaluation may be the most challenging (``adversarial'') evaluation task for multilingual text encoders as it uncovers some of their shortcomings---prominently, the inability to capture semantically non-sensical word-by-word translations or paraphrases---which remain hidden in their common evaluation scenarios. 

We release our metrics under the name \emph{XMoverScore} publicly: \url{https://github.com/AIPHES/ACL20-Reference-Free-MT-Evaluation}.   

%% file: appendix.tex
\appendix
\section{Appendix}
\subsection{Zero-shot Transfer to \Wei{Resource-lean} Language} 
\label{sec:transfer}
Our metric
%can free from human references in MT evaluation, which 
allows for estimating translation quality on new domains. However, %it still limits 
the evaluation is limited to those languages  covered by multilingual embeddings. 
%Its importance inspires us to 
%% or not fully trained (i.e., there is no large corpus for training Kazakh). 
%investigate the transferability of our metrics.
This is a major drawback for low-resource languages---e.g., Gujarati is not included in LASER. 
To this end,
% in these two scenarios: 1) 
we take multilingual USE \cite{yang2019multilingual} as an illustrating example which covers only 16 languages (in our sample Czech, Latvian and Finish are not included in USE). 
We re-align the corresponding embedding spaces with our re-mapping functions to induce evaluation metrics even for these languages, using only 2k translation pairs. 
Table \ref{tab:zero_shot_tranfer_wmt17} shows that our metric with a composition of re-mapping functions can raise correlation from zero to 0.10 for cs-en and to 0.18 for lv-en. %However, the correlation fizzles out from the negative to zero for fi-en. %\todo{SE: why?}
However, for one language pair, fi-en, we see correlation goes from negative to zero, indicating that this approach does not always work. 
% 2) Table \ref{tab:wmt19-to-en-sys} shows that our metric with original LASER performs poorly in minority languages (gu and kk), since their corresponding embeddings are not fully trained on a large corpus.\todo{SE: BERT and LASER have smal corpora for these languages?} We find that re-mapping can largely improve the correlation on kk-en from -0.611 to 0.498, but cannot contribute much on gu-en. 
%The cross-lingual alignment can advance human correlations but cannot help much on certain language pairs. 
This observation warrants further investigation.
\begin{table}[h!]
    \setlength{\tabcolsep}{4.5pt}
    \footnotesize    
    \centering
    \begin{tabular}{l | l ccccccc c}
    \toprule
    % & & \multicolumn{8}{c}{\textbf{Direct Assessment}}\\
    Metrics & cs-en & fi-en & lv-en  \\
    \midrule
    % \multirow{1}{*}{$m(\mathbf{y^*},\mathbf{y})$}
    \metric{BLEU} & 0.849 & 0.834 & 0.946 \\
    \midrule
    % \multirow{5}{*}{$m(\mathbf{x},\mathbf{y})$} 
    \metric{Cosine + LAS} & -0.001 & -0.149 & 0.019 \\
    \metric{Cosine + CLP(USE)} & 0.072 & -0.068 & 0.109 \\
    \metric{Cosine + UMD(USE)} & 0.056 & -0.061 & 0.113 \\
    \metric{Cosine + CLP $\circ$ UMD(USE)} & 0.089 & -0.030 & 0.162 \\
    \metric{Cosine + UMD $\circ$ CLP(USE)} & \textbf{0.102} & -0.007 & \textbf{0.180} \\
    % \metric{Cosine + CLP(USE) $\oplus$ LM} & - & - & - \\
    % \metric{Cosine + UMD(USE) $\oplus$ LM} & - & - & - \\
	\bottomrule  
    \end{tabular}
    \caption{The Pearson correlation of merics on segment-level WMT17. '$\circ$' marks the composition of two re-mapping functions.
    \label{tab:zero_shot_tranfer_wmt17} }
% \vspace{-0.1in}
\end{table}

\input{tables/wmt17-system}

\input{tables/wmt18}

\input{tables/wmt19}

%% file: tables/wmt17-system.tex
\begin{table*}[h!]
    \setlength{\tabcolsep}{3.3pt}
    \footnotesize    
    \centering
    \begin{tabular}{l | l ccccccc c}
    \toprule
    % & & \multicolumn{8}{c}{\textbf{Direct Assessment}}\\
    Setting & Metrics & cs-en & de-en & fi-en & lv-en & ru-en & tr-en & zh-en & Average \\
    
    \midrule
    \multirow{2}{*}{$m(\mathbf{y^*},\mathbf{y})$}
    &\metric{BLEU} & \textbf{0.971} & 0.923 & 0.903 & \textbf{0.979} & 0.912 & \textbf{0.976} & 0.864 & 0.933\\
    &\metric{chrF++} & 0.940 & \textbf{0.965} & \textbf{0.927} & 0.973 & \textbf{0.945} & 0.960 & \textbf{0.880} & \textbf{0.941}\\
    \midrule
    \multirow{17}{*}{$m(\mathbf{x},\mathbf{y})$} 
    &\multicolumn{9}{l}{\textit{Baseline with Original Embeddings}}\\
    \cmidrule{2-10}
    &\metric{Mover-1 + M-BERT} & 0.408 & \textbf{0.905} & 0.570 & 0.571 & 0.855 & 0.576 & \textbf{0.816} & 0.672\\
    &\metric{Cosine + LASER} & \textbf{0.821} & 0.821 & \textbf{0.744} & \textbf{0.754} & \textbf{0.895} & \textbf{0.890} & 0.676 & \textbf{0.800}\\
    
    \cmidrule{2-10}
    &\multicolumn{9}{l}{\textit{Cross-lingual Alignment for Sentence Embedding}}\\
    \cmidrule{2-10}
    &\metric{Cosine + CLP(LASER)} & 0.824 & 0.830 & \textbf{0.760} & \textbf{0.766} & 0.900 & \textbf{0.942} & \textbf{0.757} & \textbf{0.826}\\
    &\metric{Cosine + UMD(LASER)} & \textbf{0.833} & \textbf{0.858} & 0.735 & 0.754 & \textbf{0.909} & 0.870 & 0.630 & 0.798 \\
    % &\metric{Cosine + UMD $\circ$ CLP(LASER)} & 0.830 & 0.860 & \textbf{0.828} & \textbf{0.803} & 0.897 & \textbf{0.982} & \textbf{0.792} & \textbf{0.856}\\
    % &\metric{Cosine + CLP $\circ$ UMD(LASER)} & 0.826 & \textbf{0.869} & 0.778 & 0.766 & \textbf{0.939} & 0.950 & 0.765 & 0.842 \\
    
    \cmidrule{2-10}
    &\multicolumn{9}{l}{\textit{Cross-lingual Alignment for Word Embedding}}\\
    \cmidrule{2-10}
    &\metric{Mover-1 + RCSLS} &-0.693 & -0.053 & 0.738 & 0.251 & 0.538 & 0.380 & 0.439 & 0.229\\
    &\metric{Mover-1 + CLP(M-BERT)} & 0.796 & 0.960 & 0.879 & 0.874 & 0.894 & 0.864 & 0.898 & 0.881\\
    &\metric{Mover-2 + CLP(M-BERT)} & \textbf{0.818} & 0.971 & \textbf{0.885} & \textbf{0.887} & 0.878 & \textbf{0.893} & 0.896 & \textbf{0.890}\\
    &\metric{Mover-1 + UMD(M-BERT)} & 0.610 & 0.956 & 0.526 & 0.599 & \textbf{0.906} & 0.538 & 0.898 & 0.719\\
    &\metric{Mover-2 + UMD(M-BERT)} & 0.650 & \textbf{0.973} & 0.574 & 0.649 & 0.888 & 0.634 & \textbf{0.901} & 0.753\\
    % &\metric{Mover-1 + UMD $\circ$ CLP(M-BERT)} & 0.643 & 0.949 & 0.883 & 0.744 & 0.898 & 0.876 & 0.873 & 0.838\\
    % &\metric{Mover-1 + CLP $\circ$ UMD(M-BERT)} & 0.782 & 0.959 & 0.878 & 0.863 & 0.896 & 0.862 & 0.898 & 0.877\\
    % &\metric{Mover-2 + CLP $\circ$ UMD(M-BERT)} & 0.806 & \textbf{0.971} & \textbf{0.885} & 0.878 & 0.881 & \textbf{0.894} & 0.895 & 0.887\\
    \cmidrule{2-10}
    &\multicolumn{9}{l}{\textit{Combining Language Model}}\\
    \cmidrule{2-10}
    &\metric{Cosine + CLP(LASER) $\oplus$ LM} & \textbf{0.986} & 0.909 & 0.868 & 0.968 & 0.858 & 0.910 & 0.800 & 0.900 \\
    &\metric{Cosine + UMD(LASER) $\oplus$ LM} & 0.984 & 0.904 & 0.861 & \textbf{0.968} & 0.850 & \textbf{0.922} & \textbf{0.817} & \textbf{0.901} \\
    &\metric{Mover-2 + CLP(M-BERT) $\oplus$  LM} & 0.977 & 0.923 & \textbf{0.873} & 0.944 & 0.863 & 0.880 & 0.803 & 0.895\\
    &\metric{Mover-2 + UMD(M-BERT) $\oplus$  LM} & 0.968 & \textbf{0.934} & 0.832 & 0.951 & \textbf{0.871} & 0.862 & 0.821 & 0.891\\
    % &\metric{Cosine + UMD $\circ$ CLP(LASER) $\oplus$ LM} & 0.976 & 0.895 & 0.868 & 0.965 & 0.847 & 0.893 & 0.761 & 0.886\\
    % &\metric{Cosine + CLP $\circ$ UMD(LASER) + LM} & \textbf{0.985} & 0.910 & 0.869 & \textbf{0.968} & 0.862 & 0.911 & 0.801 & 0.901\\
    % &\metric{Mover-1 + CLP $\circ$ UMD(M-BERT) + LM} & 0.974 & \textbf{0.928} & \textbf{0.873} & 0.950 & 0.865 & 0.883 & 0.800 & 0.896 \\
    % &\metric{Mover-2 + CLP $\circ$ UMD(M-BERT) + LM} & 0.977 & 0.925 & 0.874 & 0.949 & \textbf{0.865} & \textbf{0.883} & \textbf{0.805} & \textbf{0.897} \\
	\bottomrule  
    \end{tabular}
    \caption{Pearson correlations with system-level human judgments on the WMT17 dataset. 
    \label{tab:wmt17-to-en-sys} }
% \vspace{-0.1in}
\end{table*}

%% file: tables/wmt18.tex
\begin{table*}[h!]
    \setlength{\tabcolsep}{3.3pt}
    \footnotesize    
    \centering
    \begin{tabular}{l | l ccccccc c}
    \toprule
    % & & \multicolumn{8}{c}{\textbf{Direct Assessment}}\\
    Setting & Metrics & cs-en & de-en & et-en & fi-en & ru-en & tr-en & zh-en & Average \\
    
    \midrule
    \multirow{2}{*}{$m(\mathbf{y^*},\mathbf{y})$}
    &\metric{sentBLEU} & 0.233 & 0.415 & 0.285 & 0.154 & 0.228 & 0.145 & 0.178 & 0.234 \\
    &\metric{YiSi-1} & \textbf{0.319} & \textbf{0.488} & \textbf{0.351} & \textbf{0.231} & \textbf{0.300} & \textbf{0.234} & \textbf{0.211} & \textbf{0.305} \\
    % &\metric{Meteor++} & 0.270 & 0.457 & 0.329 & 0.207 & 0.253 & 0.204 & 0.179 & 0.271 \\
    \midrule
    \multirow{17}{*}{$m(\mathbf{x},\mathbf{y})$} 
    &\multicolumn{9}{l}{\textit{Baseline with Original Embeddings}}\\
    \cmidrule{2-10}
    &\metric{Mover-1 + M-BERT} & 0.005 & 0.229 & 0.179 & 0.115 & 0.100 & 0.039 & \textbf{0.082} & 0.107 \\
    &\metric{Cosine + LASER} & \textbf{0.072} & \textbf{0.317} & \textbf{0.254} & \textbf{0.155} & \textbf{0.102} & \textbf{0.086} & 0.064 & \textbf{0.150} \\
    \cmidrule{2-10}
    &\multicolumn{9}{l}{\textit{Cross-lingual Alignment for Word Embedding}}\\
    \cmidrule{2-10}
    &\metric{Cosine + CLP(LASER)} & 0.093 & 0.323 & 0.254 & \textbf{0.151} & 0.112 & 0.086 & 0.074 & 0.156 \\
    &\metric{Cosine + UMD(LASER)} & 0.077 & 0.317 & 0.252 & \textbf{0.145} & 0.136 & 0.083 & 0.053 & 0.152 \\
    &\metric{Cosine + UMD $\circ$ CLP(LASER)} & 0.090 & \textbf{0.337} & \textbf{0.255} & 0.139 & \textbf{0.145} & \textbf{0.090} & \textbf{0.088} & \textbf{0.163} \\
    &\metric{Cosine + CLP $\circ$ UMD(LASER)} & \textbf{0.096} & 0.331 & 0.254 & 0.153 & 0.122 & 0.084 & 0.076 & 0.159 \\
    
    \cmidrule{2-10}
    &\multicolumn{9}{l}{\textit{Cross-lingual Alignment for Sentence Embedding}}\\
    \cmidrule{2-10}
    % &\metric{Mover-1 + CLP(FastText)} & - & - & - & - & - & - & - & - \\
    &\metric{Mover-1 + CLP(M-BERT)} & \textbf{0.084} & 0.279 & 0.207 & 0.147 & \textbf{0.145} & \textbf{0.089} & \textbf{0.122} & \textbf{0.153} \\
    &\metric{Mover-2 + CLP(M-BERT)} & 0.063 & \textbf{0.283} & 0.193 & 0.149 & 0.136 & 0.069 & 0.115 & 0.144 \\
    &\metric{Mover-1 + UMD(M-BERT)} & 0.043 & 0.264 & 0.193 & 0.136 & 0.138 & 0.051 & 0.113 & 0.134 \\
    &\metric{Mover-2 + UMD(M-BERT)} & 0.040 & 0.268 & 0.188 & 0.143 & 0.141 & 0.055 & 0.111 & 0.135 \\
    &\metric{Mover-1 + UMD $\circ$ CLP(M-BERT)} & 0.024 & 0.282 & 0.192 & 0.144 & 0.133 & 0.085 & 0.089 & 0.136 \\
    &\metric{Mover-1 + CLP $\circ$ UMD(M-BERT)} & 0.073 & 0.277 & \textbf{0.208} & 0.148 & 0.142 & 0.086 & 0.121 & 0.151 \\
    &\metric{Mover-2 + CLP $\circ$ UMD(M-BERT)} & 0.057 & 0.283 & 0.194 & \textbf{0.149} & 0.137 & 0.069 & 0.114 & 0.143 \\
    \cmidrule{2-10}
    &\multicolumn{9}{l}{\textit{Combining Language Model}}\\
    \cmidrule{2-10}

    &\metric{Cosine + UMD $\circ$ CLP(LASER) $\oplus$ LM} & 0.288 & 0.455 & 0.226 & 0.321 & 0.263 & 0.159 & 0.192 & 0.272\\
    &\metric{Cosine + CLP $\circ$ UMD(LASER) $\oplus$ LM} & 0.283 & 0.457 & 0.228 & 0.321 & 0.265 & 0.150 & 0.198 & 0.272\\
    &\metric{Mover-1 + CLP $\circ$ UMD(M-BERT) $\oplus$ LM} & 0.268 & 0.428 & 0.292 & 0.213 & 0.261 & 0.152 & 0.192 & 0.258 \\
    &\metric{Mover-2 + CLP $\circ$ UMD(M-BERT) $\oplus$ LM} & 0.254 & 0.426 & 0.285 & 0.203 & 0.251 & 0.146 & 0.193 & 0.251 \\
    \bottomrule  
    \end{tabular}
    \caption{Kendall correlations with segment-level human judgments on the WMT18 dataset. 
    \label{tab:wmt18-seg} }
\vspace{-0.2in}
\end{table*}

\begin{table*}[h!]
    \setlength{\tabcolsep}{3.3pt}
    \footnotesize    
    \centering
    \begin{tabular}{l | l ccccccc c}
    \toprule
    % & & \multicolumn{8}{c}{\textbf{Direct Assessment}}\\
    Setting & Metrics & cs-en & de-en & et-en & fi-en & ru-en & tr-en & zh-en & Average \\
    
    \midrule
    \multirow{2}{*}{$m(\mathbf{y^*},\mathbf{y})$}
    &\metric{BLEU} & \textbf{0.970} & 0.971 & \textbf{0.986} & \textbf{0.973} & 0.979 & 0.657 & \textbf{0.978} & 0.931 \\
    &\metric{meteor++} & 0.945 & \textbf{0.991} & 0.978 & 0.971 & \textbf{0.995} & \textbf{0.864} & 0.962 & \textbf{0.958} \\
    \midrule
    \multirow{17}{*}{$m(\mathbf{x},\mathbf{y})$} 
    &\multicolumn{9}{l}{\textit{Baseline with Original Embeddings}}\\
    \cmidrule{2-10}
    &\metric{Mover-1 + M-BERT} & -0.629 & 0.915 & 0.880 & 0.804 & 0.847 & 0.731 & \textbf{0.677} & 0.604 \\
    &\metric{Cosine + LASER} & \textbf{-0.348} & \textbf{0.932} & \textbf{0.930} & \textbf{0.906} & \textbf{0.902} & \textbf{0.832} & 0.471 & \textbf{0.661} \\
    
    \cmidrule{2-10}
    &\multicolumn{9}{l}{\textit{Cross-lingual Alignment for Sentence Embedding}}\\
    \cmidrule{2-10}
    &\metric{Cosine + CLP(LASER)} & -0.305 & 0.934 & 0.937 & 0.908 & 0.904 & 0.801 & 0.634 & 0.688 \\
    &\metric{Cosine + UMD(LASER)} & -0.241 & 0.944 & 0.933 & 0.906 & 0.902 & 0.842 & 0.359 & 0.664 \\
    &\metric{Cosine + UMD $\circ$ CLP(LASER)} & \textbf{0.195} & \textbf{0.955} & \textbf{0.958} & \textbf{0.913} & 0.896 & \textbf{0.899} & \textbf{0.784} & \textbf{0.800} \\
    &\metric{Cosine + CLP $\circ$ UMD(LASER)} & -0.252 & 0.942 & 0.941 & 0.908 & \textbf{0.919} & 0.811 & 0.642 & 0.702 \\
    
    \cmidrule{2-10}
    &\multicolumn{9}{l}{\textit{Cross-lingual Alignment for Word Embedding}}\\
    \cmidrule{2-10}
    % &\metric{Mover-1 + CLP(FastText)} & - & - & - & - & - & - & - & - \\
    &\metric{Mover-1 + CLP(M-BERT)} & \textbf{-0.163} & 0.943 & \textbf{0.918} & \textbf{0.941} & 0.915 & 0.628 & \textbf{0.875} & \textbf{0.722} \\
    &\metric{Mover-2 + CLP(M-BERT)} & -0.517 & 0.944 & 0.909 & 0.938 & 0.913 & 0.526 & 0.868 & 0.654 \\
    &\metric{Mover-1 + UMD(M-BERT)} & -0.380 & 0.927 & 0.897 & 0.886 & 0.919 & \textbf{0.679} & 0.855 & 0.683 \\
    &\metric{Mover-2 + UMD(M-BERT)} & -0.679 & 0.929 & 0.891 & 0.896 & \textbf{0.920} & 0.616 & 0.858 & 0.633 \\
    &\metric{Mover-1 + UMD $\circ$ CLP(M-BERT)} & -0.348 & \textbf{0.949} & 0.905 & 0.890 & 0.905 & 0.636 & 0.776 & 0.673 \\
    &\metric{Mover-1 + CLP $\circ$ UMD(M-BERT)} & -0.205 & 0.943 & 0.916 & 0.938 & 0.913 & 0.641 & 0.871 & 0.717 \\
    &\metric{Mover-2 + CLP $\circ$ UMD(M-BERT)} & -0.555 & 0.944 & 0.908 & 0.935 & 0.911 & 0.551 & 0.863 & 0.651 \\
    \cmidrule{2-10}
    &\multicolumn{9}{l}{\textit{Combining Language Model}}\\
    \cmidrule{2-10}
    &\metric{Cosine + UMD $\circ$ CLP(LASER) $\oplus$ LM} &0.979 & 0.967 & 0.979 & 0.947 & 0.942 & 0.673 & 0.954 & 0.919\\
    &\metric{Cosine + CLP $\circ$ UMD(LASER) $\oplus$ LM} & 0.974 & 0.966 & 0.983 & 0.951 & 0.951 & 0.255 & 0.961 & 0.863\\
    &\metric{Mover-1 + CLP $\circ$ UMD(M-BERT) $\oplus$ LM} & 0.956 & 0.960 & 0.949 & 0.973 & 0.951 & 0.097 & 0.954 & 0.834\\
    &\metric{Mover-2 + CLP $\circ$ UMD(M-BERT) $\oplus$ LM} & 0.959 & 0.961 & 0.947 & 0.979 & 0.951 & -0.036 & 0.952 & 0.815 \\
    \bottomrule  
    \end{tabular}
    \caption{Pearson correlations with system-level human judgments on the WMT18 dataset. 
    \label{tab:wmt18-sys} }
\vspace{-0.2in}
\end{table*}

%% file: tables/wmt19.tex
\begin{table*}[ht]
    \setlength{\tabcolsep}{3.3pt}
    \footnotesize    
    \centering
    \begin{tabular}{l | l ccccccc c}
    \toprule
    & & \multicolumn{8}{c}{\textbf{Direct Assessment}}\\
    Setting & Metrics & de-en & fi-en & gu-en & kk-en & lt-en & ru-en & zh-en & Average\\
    
    \midrule
    \multirow{1}{*}{$m(\mathbf{y^*},\mathbf{y})$}
    &\metric{BLEU} & 0.849 & 0.982 & 0.834 & 0.946 & 0.961 & 0.879 & 0.899 & 0.907\\
    % &\metric{meteor++} & \textbf{0.896} & \textbf{0.995} & \textbf{0.900} & \textbf{0.971} & 0.927 & \textbf{0.952} & \textbf{0.952} & \textbf{0.942}\\
    \midrule
    \multirow{17}{*}{$m(\mathbf{x},\mathbf{y})$} 
    &\multicolumn{9}{l}{\textit{Existing Reference-free Metrics}}\\
    \cmidrule{2-10}
    &\metric{ibm1-morpheme}\cite{popovic-2012-morpheme} & 0.345 & 0.740 & - & - & 0.487 & - & - & - \\
    &\metric{ibm1-pos4gram}\cite{popovic-2012-morpheme} & 0.339 & - & - & - & - & - & - & - \\
    &\metric{LASIM}\cite{yankovskaya-etal-2019-quality} & 0.247 & - & - & - & - & 0.310 & - & -\\
    &\metric{LP}\cite{yankovskaya-etal-2019-quality} & 0.474 & - & - & - & - & 0.488 & - & -\\
    % &\metric{UNI} & 0.846 & 0.93 & - & - & - & 0.805 & - & -\\
    % &\metric{UNI+} & 0.850 & 0.924 & - & - & - & 0.808 & - & - \\
    &\metric{YiSi-2}\cite{lo-2019-yisi} & 0.796 & 0.642 & 0.566 & 0.324 & 0.442 & 0.339 & 0.940 & 0.578\\
    &\metric{YiSi-2-srl}\cite{lo-2019-yisi} & 0.804 & - & - & - & - & - & 0.947 & -\\
    \cmidrule{2-10}
    
    &\multicolumn{9}{l}{\textit{Baseline with Original Embeddings}}\\
    \cmidrule{2-10}
    &\metric{Mover-1 + M-BERT} & \textbf{0.358} & 0.611 & \textbf{-0.396} & \textbf{0.335} & \textbf{0.559} & \textbf{0.261} & \textbf{0.880} & \textbf{0.373}\\
    &\metric{Cosine + LASER} & 0.217 & \textbf{0.891} & -0.745 & -0.611 & 0.683 & -0.303 & 0.842 & 0.139\\
    \cmidrule{2-10}
    &\multicolumn{9}{l}{\textit{Our Cross-lingual based Metrics}}\\
    \cmidrule{2-10}
    &\metric{Mover-2 + CLP(M-BERT)} & \textbf{0.625} & 0.890 & -0.060 & \textbf{0.993} & \textbf{0.851} & \textbf{0.928} & \textbf{0.968} & \textbf{0.742}\\
    &\metric{Cosine + CLP(LASER)} & 0.225 & \textbf{0.894} & \textbf{0.041} & 0.150 & 0.696 & -0.184 & 0.845 & 0.381\\
    &\metric{Cosine + UMD $\circ$ CLP(LASER)} & 0.074 & 0.835 & -0.633 & \textbf{0.498} & \textbf{0.759} & -0.201 & 0.610 & 0.277\\

    \cmidrule{2-10}
    &\multicolumn{9}{l}{\textit{Our Cross-lingual based Metrics $\oplus$ LM}}\\
    \cmidrule{2-10}
    &\metric{Cosine + CLP(LASER) $\oplus$ LM} & 0.813 & 0.910 & -0.070 & -0.735 & 0.931 & 0.630 & 0.711 & 0.456\\
    &\metric{Cosine + UMD(LASER) $\oplus$ LM} & 0.817 & 0.908 & -0.383 & -0.902 & 0.929 & 0.573 & 0.781 & 0.389\\
    &\metric{Mover-2 + CLP(M-BERT) $\oplus$  LM} & 0.848 & 0.907 & \textbf{-0.068} & \textbf{0.775} & 0.963 & \textbf{0.866} & 0.827 & \textbf{0.731}\\
    &\metric{Mover-2 + UMD(M-BERT) $\oplus$  LM} & \textbf{0.859} & \textbf{0.914} & -0.181 & \textbf{-0.391} & \textbf{0.970} & 0.702 & \textbf{0.874} & 0.535\\
    % &\metric{Cosine + UMD $\circ$ CLP(LASER) + LM} & 0.809 & 0.898 & -0.151 & -0.661 & 0.935 & 0.613 & 0.589 & 0.433\\
    % &\metric{Cosine + CLP $\circ$ UMD(LASER) + LM} & 0.817 & 0.909 & -0.074 & -0.729 & 0.931 & 0.629 & 0.716 & 0.457\\
    % &\metric{Mover-1 + CLP $\circ$ UMD(M-BERT) + LM} & 0.847 & 0.908 & -0.093 & 0.704 & 0.954 & 0.817 & 0.813 & 0.707\\
    % &\metric{Mover-2 + CLP $\circ$ UMD(M-BERT) + LM} & 0.849 & 0.907 & -0.064 & 0.792 & 0.964 & 0.864 & 0.833 & 0.735 \\
    \bottomrule  
    \end{tabular}
    \caption{Pearson correlations with system-level human judgments on the WMT19 dataset. '-' marks the numbers not officially reported in \cite{ma-etal-2019-results}. 
    \label{tab:wmt19-to-en-sys} 
    }
% \vspace{-0.1in}
\end{table*}